\documentclass{article}

\usepackage{microtype}
\usepackage{graphicx}
\usepackage{subfigure}
\usepackage{booktabs}
\usepackage{stfloats}
\usepackage{makecell}
\usepackage{longtable}

\usepackage{hyperref}

\usepackage[accepted]{icml2023}  

\usepackage{amsmath}
\usepackage{amssymb}
\usepackage{mathtools}
\usepackage{amsthm}
\usepackage{wasysym}  

\usepackage[framemethod=tikz]{mdframed}  

\usepackage[capitalize,noabbrev]{cleveref}

\usepackage{lscape}

\theoremstyle{plain}

\theoremstyle{definition}

\theoremstyle{remark}

\usepackage{enumitem}
\setlist{nolistsep}

\usepackage{xcolor}
\usepackage{soul}
\usepackage{colortbl}
\definecolor{refinedweb}{HTML}{DB57B2}
\definecolor{rw_filtered}{HTML}{B55DD4}
\definecolor{rw_raw}{HTML}{5E57D3}
\definecolor{openai}{HTML}{5F57DB}
\definecolor{opt}{HTML}{DBA157}
\definecolor{eai}{HTML}{57DBC2}
\definecolor{bs}{HTML}{92DB57}
\definecolor{aleph}{HTML}{D3DB57}
\definecolor{fairscale}{HTML}{DBC257}
\definecolor{palm}{HTML}{DB5F56}
\definecolor{cerebras}{HTML}{57D3DB}
\definecolor{pile}{HTML}{57DB5F}
\definecolor{pythia}{HTML}{5691DB}
\definecolor{pile}{HTML}{7DD86E}

\icmltitlerunning{The RefinedWeb dataset for Falcon LLM}

\begin{document}
\setlength{\LTcapwidth}{\textwidth}
\onecolumn

\icmltitle{The RefinedWeb Dataset for Falcon LLM:\texorpdfstring{\\}{ }Outperforming Curated Corpora with Web Data, and Web Data Only}

\begin{icmlauthorlist}
\underline{\textbf{The Falcon LLM team}} \vspace{0.1in} \\
\icmlauthor{Guilherme Penedo}{lighton}
\icmlauthor{Quentin Malartic}{tii} \\
\icmlauthor{Daniel Hesslow}{lighton}
\icmlauthor{Ruxandra Cojocaru}{tii}
\icmlauthor{Alessandro Cappelli}{lighton}
\icmlauthor{Hamza Alobeidli}{tii}
\icmlauthor{Baptiste Pannier}{lighton} \\
\icmlauthor{Ebtesam Almazrouei}{tii}
\icmlauthor{Julien Launay}{lighton,ens}
\end{icmlauthorlist}
\icmlaffiliation{lighton}{LightOn}
\icmlaffiliation{tii}{Technology Innovation Institute, 9639 Masdar City, Abu Dhabi, United Arab Emirates}
\icmlaffiliation{ens}{LPENS, École normale supérieure}
\icmlcorrespondingauthor{}{falconllm@tii.ae} 

\icmlkeywords{deduplication, NLP, LLM, curated, data, crawl, Falcon LLM}
\printAffiliationsAndNotice{}  

\vspace{0in}

\begin{center}
\url{https://huggingface.co/datasets/tiiuae/falcon-refinedweb}
\vskip 0.2in
\end{center}

\begin{abstract}
Large language models are commonly trained on a mixture of filtered web data and curated ``high-quality'' corpora, such as social media conversations, books, or technical papers. This curation process is believed to be necessary to produce performant models with broad zero-shot generalization abilities. However, as larger models requiring pretraining on trillions of tokens are considered, it is unclear how scalable is curation and whether we will run out of unique high-quality data soon.  At variance with previous beliefs, we show that properly filtered and deduplicated web data alone can lead to powerful models; even significantly outperforming models from the state-of-the-art trained on The Pile. Despite extensive filtering, the high-quality data we extract from the web is still plentiful, and we are able to obtain five trillion tokens from CommonCrawl. We publicly release an extract of 600 billion tokens from our \textsc{RefinedWeb} dataset, and 1.3/7.5B parameters language models trained on it\footnote{Details about how to access Falcon LLM open source is available on \url{falconllm.tii.ae}}. 
\end{abstract}

\begin{figure}[h]
\centering
\includegraphics[width=.50\textwidth]{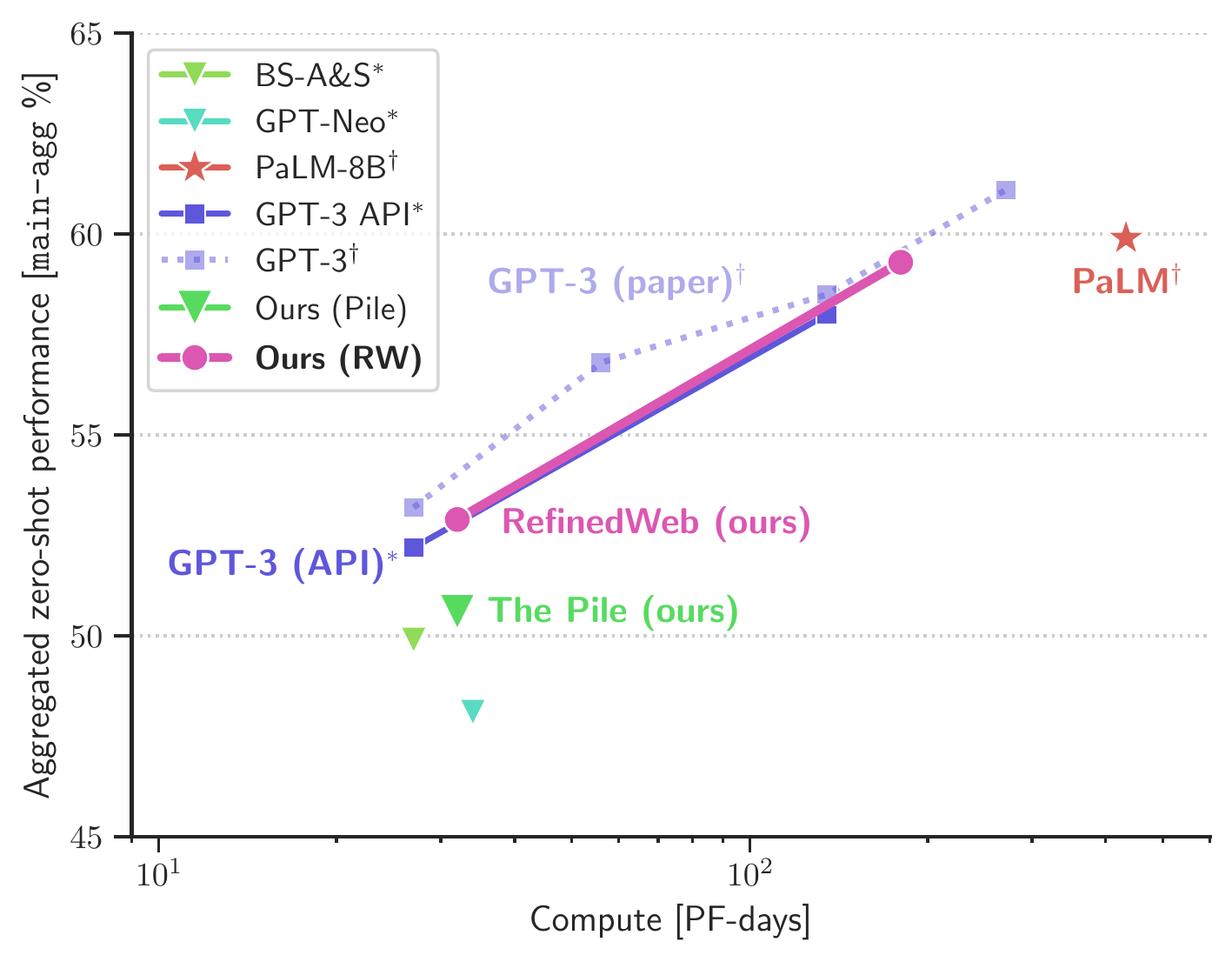}
\caption{\textbf{Models trained on \textcolor{refinedweb}{\CIRCLE \textsc{RefinedWeb}} alone outperform models trained on curated corpora.} Zero-shot performance on our~\texttt{main-agg} task aggregate (see \cref{sec:exp_setting} for details). At equivalent compute budgets, our models significantly outperform publicly available models trained on \textcolor{pile}{$\blacktriangledown$ The Pile}, and match the performance of the \textcolor{openai}{$\blacksquare$ GPT-3} models when tested within our evaluation setup.}
\label{fig:main_lead}
\end{figure}

\twocolumn
\newpage

\begin{table*}[t]
\centering
\caption{\textbf{\textcolor{refinedweb}{\CIRCLE \textsc{RefinedWeb}} improves on existing English pretraining datasets for large language models by combining extensive filtering with stringent deduplication at unprecedented scale.} For additional details, see the full version in \cref{tab:full_datasets} of \cref{sec:other_datasets}. }
\vspace{0.1in}
\label{tab:partial_datasets}
\begin{small}
\begin{tabular}{p{2cm}cccp{4cm}p{4.5cm}}
\toprule
\textbf{Dataset} & \textbf{Size} & \textbf{Availability} & \textbf{Web} & \textbf{CC Processing} & \textbf{Deduplication} \\ 
\midrule
\multicolumn{6}{c}{\textbf{\textsc{Massive web datasets}}} \\ \midrule
\textbf{C4} & $\sim 360$GT & Public & $100$\% & Rules + NSFW words blocklist & \textbf{Exact:} spans of 3 sentences \\
\textbf{OSCAR-21.09}  & $\sim 370$GT & Public & $100$\% & Built at the line-level & \textbf{Exact}: per line ($\sim 55\%$ removed)\\
\textbf{OSCAR-22.01} & $\sim 283$GT & Public & $100$\% & Line-level rules + optional rules \& NSFW URL blocklist &  \textbf{Exact}: per line (optional, not used for results in this paper)\\ \midrule
\multicolumn{6}{c}{\textbf{\textsc{Curated datasets}}} \\ \midrule
\textbf{\textcolor{openai}{$\blacksquare$ GPT-3}} & $300$GT & Private & $60$\% & Content filter trained on known high-quality sources & \textbf{Fuzzy}: MinHash ($\sim 10\%$ removed)\\
\textbf{\textcolor{pile}{$\blacktriangledown$ The Pile}} & $\sim 340$GT & Public & $18$\% & \texttt{jusText} for extraction, content filter trained on curated data & \textbf{Fuzzy}: MinHash ($\sim 26\%$ removed)\\ 
\textbf{\textcolor{palm}{$\bigstar$ PaLM}} & $780$GT & Private & $27$\% & Filter trained on HQ data & Unknown\\
\midrule
\multicolumn{6}{c}{\textbf{\textsc{Ours}}} \\ \midrule
\textbf{\textcolor{refinedweb}{\CIRCLE \textsc{RefinedWeb}}} & $\sim5,000$GT & Public (600GT) & $100\%$ & \texttt{trafilatura} for text extraction, document and line-level rules, NSFW URL blocklist & \textbf{Exact \& fuzzy}: exact substring+MinHash ($\sim 50\%$ removed)  \\
\bottomrule
\end{tabular}
\end{small}
\end{table*}

\section{Introduction}
\label{sec:introduction}

Progress in natural language processing is increasingly driven by sheer compute scale alone~\cite{sevilla2022compute}: as more compute is expended to train large language models~(LLM), they gain and exhibit powerful emergent capabilities~\cite{brown2020language, weiemergent}. To best benefit from scaling, recent scaling laws dictate that both model size and dataset size should jointly be increased \cite{hoffmann2022training}. This is at variance with earlier findings, which had argued that scaling should focus on model size first and foremost, with minimal data scaling \cite{kaplan2020scaling}.

This joint scaling paradigm raises significant challenges: although plentiful, text data is not infinite, especially so when considerations on data quality and licensing are taken into account--leading some researchers to argue scaling may soon be bottlenecked by data availability \cite{villalobos2022will}. Concretely, optimally training a GPT-3 sized model (175B parameters) would require no less than 3,500 billion tokens of text according to~\citet{hoffmann2022training}. This is twice as much as the largest pretraining datasets ever demonstrated \cite{hoffmann2022training, touvron2023llama}, and ten times more than the largest publicly available English datasets such as OSCAR~\cite{OrtizSuarezSagotRomary2019}, C4~\cite{2020t5}, or The Pile \cite{gao2020pile}.

Massively scaling-up pretraining data is made even more challenging by the fact LLMs are commonly trained using a mixture of web crawls and so-called ``high-quality'' data~\cite{brown2020language, gao2020pile}. Typical high-quality corpora include curated sources of books, technical documents, human-selected web pages, or social media conversations. The increased diversity and quality brought forth by these curated corpora is believed to be a key component of performant models \cite{scao2022language}. Unfortunately, curation is labour intensive: typically, each source requires specialized processing, while yielding a limited amount of data. Furthermore, licensed sources raise legal challenges.

Nevertheless, most pretraining data is still sourced from massive web crawls which can be scaled up to trillions of tokens with limited human intervention. However, the quality of this data has traditionally been seen as (much) inferior to that of the manually curated data sources. Even finely processed sources of web data, such as C4 \cite{2020t5} or OSCAR \cite{OrtizSuarezSagotRomary2019}, are regarded as inferior to curated corpora for LLMs \cite{gopher, scao2022language}, producing less performant models.

To sustain the ever-increasing data needs of larger and larger LLMs, and to streamline data pipelines and reduce the need for human-intensive curation, we propose to explore how web data can be better processed to significantly improve its quality, resulting in models as capable, if not more capable, than models trained on curated corpora. 

\vspace{-0.1in}

\paragraph{Contributions.} We make the following contributions:
\begin{itemize}
    \item We introduce \textcolor{refinedweb}{\textbf{\textsc{RefinedWeb}}}, a high-quality five trillion tokens web-only English pretraining dataset;
    \item We demonstrate that \textbf{web data alone can result in models outperforming both public and private curated corpora}, as captured by zero-shot benchmarks, challenging current views about data quality;
    \item \textbf{We publicly release a 600B tokens extract of RefinedWeb, and 1/7B parameters LLMs trained on it}, to serve as a new baseline high-quality web dataset for the natural language processing community.
\end{itemize}

\section{Related works}

\paragraph{Pretraining data for large language models.} Early large language models identified the importance of datasets with long, coherent documents \cite{radford2018improving, devlin2019bert}. Moving on from the previously used sentence-wise datasets \cite{chelba2013one}, they instead leveraged document-focused, single-domain corpora like Wikipedia or BookCorpus \cite{zhu2015aligning}. As models increased in scale, datasets based on massive web-scrape gained prevalence~\cite{OrtizSuarezSagotRomary2019, 2020t5}. However, further work argued that these untargeted web scrape fell short of human-curated data \cite{radford2019language}, leading to the wide adoption of curated datasets such as The Pile \cite{gao2020pile}, which combine web data with books, technical articles, and social media conversations. At scale, it has been proposed to emulate the human curation process by leveraging weak signals: for instance, by crawling the top links of a forum~\cite{Gokaslan2019OpenWeb}. Targeted corpora can also produce domain-specific models \cite{beltagy2019scibert}, or broaden the expressiveness of models (e.g., for conversational modalities \citet{adiwardana2020towards, thoppilan2022lamda}). Latest large language models \cite{brown2020language, gopher, chowdhery2022palm, scao2022bloom} are trained on giant aggregated corpora, combining both massive web-scrape and so-called ``high-quality'' curated single-domain sources (e.g., news, books, technical papers, social media conversations). These targeted sources are often upsampled--from one to five times is most common--to increase their representation in the final dataset. The diversity and ``higher-quality'' brought fourth by these aggregated datasets is thought to be central to model quality; web data alone is considered insufficient to train powerful large language models \cite{liu2019roberta, scao2022language}.

\paragraph{Pipelines for web data.} Massive web datasets are typically built upon CommonCrawl, a publicly available scrape of the internet, which has now been running for 12 years and has collected petabytes of data. Working with data scraped from all over the internet presents unique challenges: notably, a significant portion is low-quality machine-generated spam or pornographic content \cite{trinh2018simple, kreutzer2022quality}. Accordingly, training on unfiltered web data is undesirable, resulting in poorly performing models~\cite{2020t5}. Modern pipelines focus on filtering out this undesirable content \cite{wenzek2020ccnet}. Broadly speaking, these pipelines usually combine a variety of stages: (1) \emph{language identification}, leveraging inexpensive n-gram models (e.g., fastText \citet{joulin2016fasttext}); (2)~\emph{filtering rules and heuristics}, such as only keeping lines with valid punctuation, discarding lines with too many symbols, or removing documents containing banned words \cite{grave2018learning, 2020t5}; (3) \emph{ML-based quality filtering}, using lightweight models trained on known gold data to identify similar high-quality web documents \cite{wenzek2020ccnet, brown2020language}; (4) \emph{deduplication}, removing either exact duplicate spans or similar documents~\cite{lee2022deduplicating}. While some filtering is necessary, excessive filtering can introduce undesirable biases in the model. This can overly impact minorities~\cite{dodge2021documenting}, motivating the adoption of practices such as pseudo-crawling, wherein allowed URLs are manually curated~\cite{laurencconbigscience}.

\paragraph{Deduplication.} Deduplication removes repeated extracts and documents from a dataset: these could either be exact matches, identical in every character, or approximate matches, based on some similarity metric. For exact duplicates, it is common to match exact substrings of a minimum length using suffix arrays \cite{manber1993suffix}. For fuzzy duplicates, methods based on locally-sensitive hashes such as MinHash \cite{broder1997resemblance} or SimHash \cite{charikar2002similarity} have been adopted for the pretraining data of large language models \cite{brown2020language, zeng2021pangu, gopher}. Recently, \citet{abbas2023semdedup} has proposed to leverage embeddings from pretrained models to imbue semantic understanding in approximate matching algorithms.  Deduplication has been identified as playing a significant role in improving language models~\cite{allamanis2019adverse, lee2022deduplicating}. Notably, it reduces memorization~\cite{carlini2022quantifying}, which is especially problematic in large models \cite{carlini2021extracting}. Furthermore, repeated data has been shown to be increasingly harmful to model quality as parameter count increases~\cite{hernandez2022scaling}: for a 1B parameters model, a hundred duplicates are harmful; at 175B, even a few duplicates could have a disproportionate effect. Concurrently to this work, the Pythia suite of models found that deduplicating The Pile had a limited impact on zero-shot performance \cite{biderman2023pythia}, questioning whether deduplication is as relevant for curated corpora as it for predominantly web-based datasets.

We provide an overview of some widely adopted existing pretraining English datasets for LLMs in \cref{tab:partial_datasets}, with additional information in \cref{tab:full_datasets} of \cref{sec:other_datasets}. We also note that recent popular open models \cite{zhang2022opt, touvron2023llama} often indirectly leverage The Pile \cite{gao2020pile} by doing a mix-and-match of its components.

Focusing on building a large-scale high-quality web pretraining dataset, we extend upon the state-of-the-art in three ways: (1) we aggregate and combine best-practices for document preparation and filtering across multiple pipelines, and introduce line-wise corrections; (2) we combine both exact and fuzzy deduplication at very large-scale; (3) the scale of our final dataset is unique, with a total 5,000 billion tokens, and a 600 billion tokens extract available for public use with permissive licensing. Training large models on RefinedWeb also lead us to challenge the commonly held belief that web data is strictly worse than curated corpora.

\section{Macrodata Refinement and RefinedWeb}
\label{sec:mdr}

We introduce \textbf{\textsc{MDR}} (MacroData Refinement), a pipeline for filtering and deduplicating web data from CommonCrawl at very large scale. Using MDR, we produce  \textcolor{refinedweb}{\textbf{\textsc{RefinedWeb}}}, an English pretraining dataset of five trillion tokens based on web data only. We leverage strict filtering and stringent deduplication to uplift the quality of web data, distilling it down to a corpus matching the quality of aggregated corpora used to train state-of-the-art models.

\vspace{-0.1in}

\paragraph{Design principles.} We abide by the following guidelines:
\begin{itemize}
    \item \textbf{Scale first.} We intend MDR to produce datasets to be used to train 40-200B parameters models, thus requiring trillions of tokens \cite{hoffmann2022training}. For English-only RefinedWeb, we target a size of 3-6 trillion tokens. Specifically, we eschew any labour intensive human curation process, and focus on CommonCrawl instead of disparate single-domain sources. 
    \item \textbf{Strict deduplication.} Inspired by the work of \citet{lee2022deduplicating}, which demonstrated the value of deduplication for large language models, we implement a rigorous deduplication pipeline. We combine both exact and fuzzy deduplication, and use strict settings leading to removal rates far higher than others have reported.
    \item \textbf{Neutral filtering.} To avoid introducing further undesirable biases into the model \cite{dodge2021documenting, welbl2021challenges}, we avoid using ML-based filtering outside of language identification. We stick to simple rules and heuristics, and use only URL filtering for adult content. 
\end{itemize}

\begin{figure*}[b]
\centering
\includegraphics[width=0.76\linewidth]{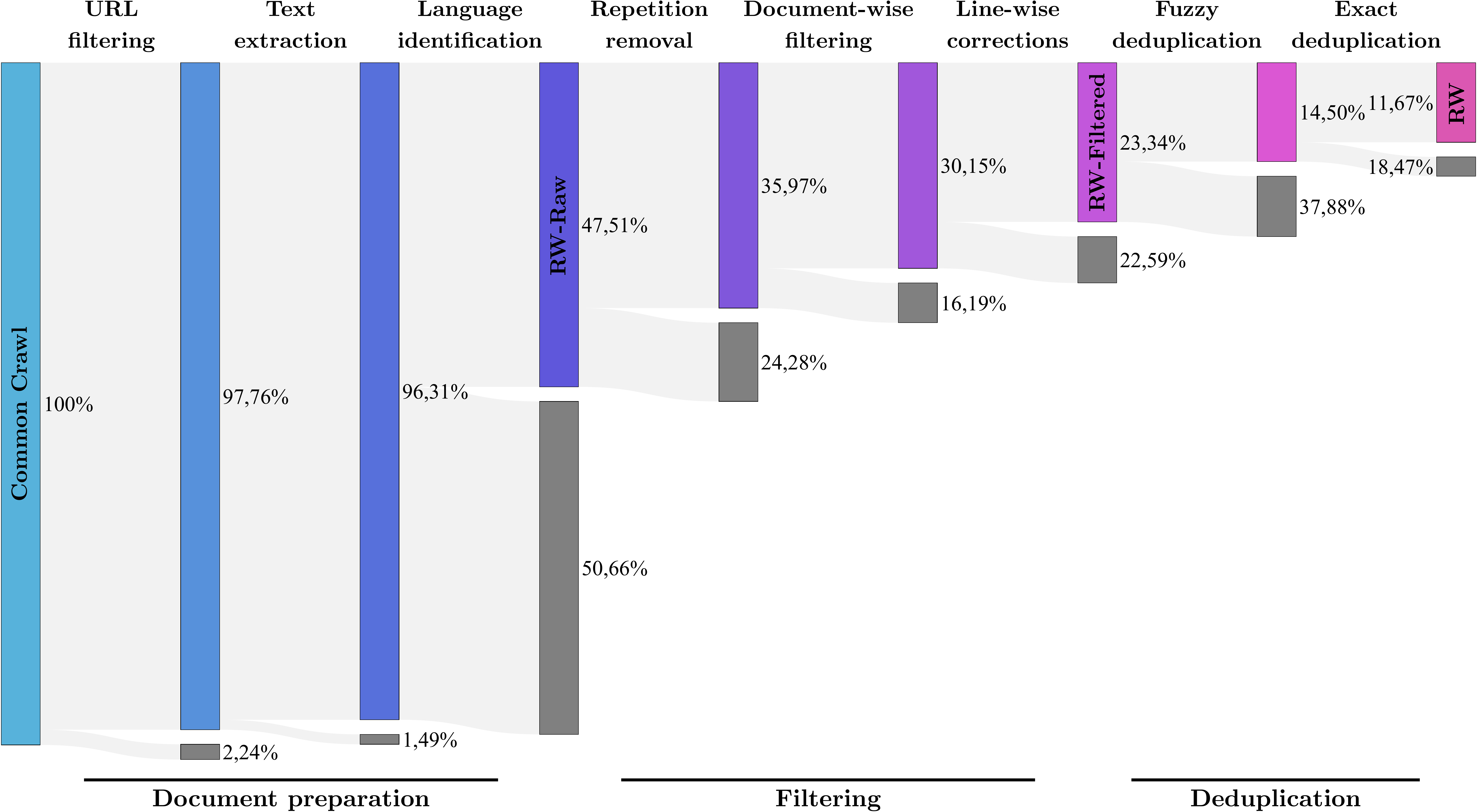}
\caption{\textbf{Subsequent stages of Macrodata Refinement remove nearly 90\% of the documents originally in CommonCrawl.} Notably, filtering and deduplication each result in a halving of the data available: around 50\% of documents are discarded for not being English, 24\% of remaining for being of insufficient quality, and 12\% for being duplicates. We report removal rate (\textcolor{gray}{grey}) with respect to each previous stage, and kept rate (\textcolor{refinedweb}{shade}) overall. Rates measured in \% of documents in the document preparation phase, then in tokens.}
\label{fig:mdr_pipeline}
\end{figure*}

\cref{tab:mdr_pipeline} and \cref{fig:mdr_pipeline} outline the full MDR pipeline.

\subsection{Document preparation: reading data, filtering URLs, extracting text, and language identification}

\paragraph{Reading the data.} CommonCrawl is available in either WARC (raw HTML response), or WET files (preprocessed to only include plain text). Individual files correspond to a page at a given URL; these constitute single documents/samples. Working with WET files would spare us from running our own HTML extraction; however, in line with previous works \cite{gao2020pile, gopher}, we found WET files to include undesirable navigation menus, ads, and other irrelevant texts. Accordingly, our pipeline starts from raw WARC files, read with the \texttt{warcio} library. 

\paragraph{URL filtering.} \label{sec:URLFiltering} Before undertaking any compute-heavy processing, we perform a first filtering based on the URL alone. This targets fraudulent and/or adult websites (e.g., predominantly pornographic, violent, related to gambling, etc.). We base our filtering on two rules: (1) an aggregated blocklist of 4.6M domains; (2) a URL score, based on the presence of words from a list we curated and weighed by severity. We found that commonly used blocklists include many false positives, such as popular blogging platforms or even pop culture websites. Furthermore, word-based rules (like the one used in C4, \citet{2020t5}) can easily result in medical and legal pages being blocked. Our final detailed rules based on this investigation are shared in \cref{sec:url_details}. Since we intend RefinedWeb to be used as part of an aggregate dataset along with curated corpora, we also filtered common sources of high-quality data: Wikipedia, arXiv, etc. The detailed list is available in \cref{sec:excluded_sources}.

\paragraph{Text extraction.} We want to extract only the main content of the page, ignoring menus, headers, footers, and ads among others: \citet{lopukhin2019} found that \texttt{trafilatura} \cite{barbaresi-2021-trafilatura} was the best non-commercial library for retrieving content from blog posts and news articles. Although this is only a narrow subset of the kind of pages making up CommonCrawl, we found this finding to hold more broadly. We  use \texttt{trafilatura} for text extraction, and apply extra formatting via regular expressions: we limit new lines to two consecutive ones, and remove all URLs. 

\paragraph{Language identification.} We use the fastText language classifier of CCNet \cite{wenzek2020ccnet} at the document-level: it uses characters n-gram and was trained on Wikipedia, supporting 176 languages. We remove documents for which the top language scores below 0.65: this usually corresponds to pages without any natural text. For this paper, we focus on English; RefinedWeb can also be derived for other languages, see \cref{sec:multilingual} for details.

The data we retrieve at this stage, called \textcolor{rw_raw}{\textsc{\textbf{RW-Raw}}}, corresponds to what we can extract with the minimal amount of filtering. At this stage, only 48\% of the original documents are left, mostly filtered out by language identification.

\begin{table*}[t]
    \centering
    \caption{\textbf{Macrodata Refinement aggregates best practices from the state-of-the-art and novel approaches (URL scoring, line-wise filtering, etc.) to produce high-quality web data.} On deduplication, we note that MDR is unique in both the scale at which it is performed, and in applying subsequently fuzzy and exact substring methods to improve coverage and scalability.}
    \label{tab:mdr_pipeline}
    \vspace{0.1in}
    \begin{scriptsize}
    \begin{tabular}{p{2cm}p{2cm}p{2cm}p{2cm}p{2cm}p{2cm}p{2cm}p{2cm}p{2cm}p{2cm}}
    \toprule
    \multicolumn{3}{l}{\textbf{\textsc{Document preparation}}} & \multicolumn{2}{l}{\textbf{\textsc{Filtering}}} & \multicolumn{2}{l}{\textbf{\textsc{Deduplication}}} \\
    \midrule
    \textbf{URL filtering} & \textbf{Text extraction} & \makecell[tl]{\textbf{Language} \\ \textbf{identification}} & \makecell[tl]{\textbf{Document-wise} \\ \textbf{filtering}} & \makecell[tl]{\textbf{Line-wise} \\ \textbf{filtering}} & \textbf{Deduplication} & \textbf{URL deduplication} \\
    \midrule
    Aggregated blocklist, URL scoring, common HQ sources blocked & From WARC using \texttt{warcio}, \texttt{trafilatura} for extraction & \texttt{fastText} classifier from CCNet, thresholding on top language score & In-document repetition removal and quality heuristics from MassiveWeb & Remove undesirable lines (call to actions, navigation buttons, social counters, etc.) & Fuzzy deduplication w/ MinHash + exact substring deduplication w/ suffix arrays & Remove URLs revisited across CommonCrawl dumps  \\
    \cref{sec:url_details} & \citet{barbaresi-2021-trafilatura} & \citet{wenzek2020ccnet} & \citet{gopher} & \cref{sec:line_details} & \citet{lee2022deduplicating} & \cref{sec:dedup_mdr} \\ \bottomrule
    \end{tabular}
    \end{scriptsize}
\end{table*}

\subsection{Filtering: document-wise and line-wise}

\paragraph{Repetition removal.} Due to crawling errors and low-quality sources, many documents contain repeated sequences: this may cause pathological behavior in the final model~\cite{holtzman2019curious}. We could catch this content at the later deduplication stage, but it is cheaper and easier to catch it document-wise early on. We implement the heuristics of \citet{gopher}, and remove any document with excessive line, paragraph, or n-gram repetitions.

\paragraph{Document-wise filtering.} A significant fraction of pages are machine-generated spam, made predominantly of lists of keywords, boilerplate text, or sequences of special characters. Such documents are not suitable for language modeling; to filter them out, we adopt the quality filtering heuristics of \citet{gopher}. These focus on removing outliers in terms of overall length, symbol-to-word ratio, and other criteria ensuring the document is actual natural language. We note that these filters have to be adapted on a per language basis, as they may result in overfiltering if naively transferred from English to other languages. 

\paragraph{Line-wise corrections.} Despite the improvements brought forth by using \texttt{trafilatura} instead of relying on preprocessed files, many documents remain interlaced with undesirable lines (e.g., social media counters {\small \texttt{3 likes}}, navigation buttons). Accordingly, we devised a line-correction filter, targeting these undesirable items. If these corrections remove more than 5\% of a document, we remove it entirely.  See \cref{sec:line_details} for details.

The data we retrieve at this stage has gone through all of the filtering heuristics in the MDR pipeline. We refer to this dataset as \textcolor{rw_filtered}{\textsc{\textbf{RW-Filtered}}}. Only 23\% of the documents of CommonCrawl are left, with around 50\% of the documents of RW-Raw removed by the filtering.

\subsection{Deduplication: fuzzy, exact, and across dumps}
\label{sec:dedup_mdr}

\begin{table*}[t]
    \centering
    \caption{\textbf{To evaluate models trained on RefinedWeb and compare to the state-of-the-art, we build four aggregates across 18 tasks on which to measure zero-shot performance.} \texttt{small} was built for internal ablations, based on tasks with consistent performance at small scale, \texttt{core} is based on tasks commonly reported for public suites of models \cite{dey2023cerebras, biderman2023pythia}, \texttt{main} is based on tasks from the GPT-3 and PaLM paper \cite{brown2020language, chowdhery2022palm}, and \texttt{ext} is based on tasks used by the BigScience Architecture and Scaling group \cite{scao2022language}. For all results reported, we flag with $\dagger$ results obtained in an arbitrary evaluation setup, and with $*$ results obtained with the EAI Harness \cite{gao2021eval}, which we also employ for all our models.}
    \vspace{0.1in}
    \label{tab:task_aggregates}
    \begin{scriptsize}
    \begin{tabular}{llccccc}
    \toprule
        \textbf{Tasks} & \textbf{Type} & \textbf{Random} & \texttt{small} & \texttt{core} & \texttt{main} & \texttt{ext} \\
        \midrule
         HellaSwag \cite{zellers2019hellaswag} & Sentence completion & 25.0 & \checkmark  & \checkmark & \checkmark & \checkmark\\
         LAMBADA \cite{paperno2016lambada} & Sentence completion & 0.0 &  & \checkmark & \checkmark & \checkmark\\
         Winogrande \cite{sakaguchi2021winogrande} & Coreference resolution & 50.0 & \checkmark & \checkmark & \checkmark & \checkmark\\ 
         PIQA \cite{bisk2020piqa} & Multiple-choice question answering & 50.0 & \checkmark & \checkmark& \checkmark & \checkmark\\ 
         ARC \cite{clark2018think} & Natural language inference & 25.0 & \checkmark & \checkmark& \checkmark & \checkmark\\ 
         OpenBookQA \cite{mihaylov2018can} & Multiple-choice question answering & 25.0 & & \checkmark& \checkmark & \checkmark\\ 
         BoolQ \cite{clark2019boolq} & Multiple-choice question answering & 50.0 & \checkmark & & \checkmark & \checkmark \\
         COPA \cite{gordon2012semeval} & Sentence completion & 50.0 & & & \checkmark & \checkmark \\ 
         CB \cite{de2019commitmentbank} & Natural language inference & 33.3 & & & \checkmark & \checkmark \\
         RTE \cite{dagan2010recognizing} & Natural language inference & 50.0 & & & \checkmark & \checkmark \\ 
         ReCoRD \cite{zhang2018record} & Question answering & 0.0 & & & \checkmark & \\ 
         ANLI \cite{nie2019adversarial} & Natural language inference & 33.3 & & & \checkmark &  \\ 
         LogiQA \cite{liu2021logiqa} & Multiple-choice question answering & 25.0 & & & & \checkmark \\
         HeadQA \cite{vilares2019head} & Multiple-choice question answering & 20.0 & & & & \checkmark \\ 
         MathQA \cite{amini2019mathqa}  &Multiple-choice question answering & 20.0 & & & & \checkmark \\ 
         PROST \cite{aroca2021prost} & Paraphrase identification & 50.0 & & & & \checkmark \\ 
         PubMedQA \cite{jin2019pubmedqa} & Multiple-choice question answering & 50.0 & & & & \checkmark \\
         SciQ \cite{welbl2017crowdsourcing} & Multiple-choice question answering & 25.0 & \checkmark & & & \checkmark \\ 
        \bottomrule
    \end{tabular}
    \end{scriptsize}
        \vspace{-0.15in}
\end{table*}

After filtering, although data quality has improved, a large fraction of the content is repeated across documents. This may be due to the crawler indirectly hitting the same page multiple times, to boilerplate content being repeated (e.g., licences), or even to plagiarism. These duplicates can strongly impact models, favoring memorization instead of generalization~\cite{lee2022deduplicating, hernandez2022scaling}. Since deduplication is expensive, it has seen limited adoption in public datasets \cite{OrtizSuarezSagotRomary2019, 2020t5}. We adopt an aggressive deduplication strategy, combining both fuzzy document matches and exact sequences removal.

\paragraph{Fuzzy deduplication.} We remove similar documents by applying MinHash \cite{broder1997resemblance}: for each document, we compute a sketch and measure its approximate similarity with other documents, eventually removing pairs with high overlap. MinHash excels at finding templated documents: licenses with only specific entities differing, placeholder SEO text repeated across websites--see examples of the biggest clusters in \cref{sec:minhash_cluster}. We perform MinHash deduplication using 9,000 hashes per document, calculated over 5-grams and divided into 20 buckets of 450 hashes. We found that using less aggressive settings, such as the 10 hashes of The Pile \cite{gao2020pile}, resulted in lower deduplication rates and worsened model performance. See \cref{sec:minhash_details} for more details about our MinHash setup.  
\vspace{-0.1in}

\paragraph{Exact deduplication.} Exact substring operates at the sequence-level instead of the document-level, finding matches between strings that are exact token-by-token matches by using a suffix array \cite{manber1993suffix} (e.g., specific disclaimers or notices, which may not compromise the entire document as showcased in \cref{sec:exact_matches}). We remove any match of more than 50 consecutive tokens, using the implementation of \citet{lee2022deduplicating}. We note that exact substring alters documents, by removing specific spans: we also experimented with dropping entire documents or loss-masking the duplicated strings instead of cutting them, but this didn't result in significant changes in zero-shot performance--see \cref{sec:exact_details}. 
\vspace{-0.1in}

\paragraph{URL deduplication.} Because of computational constraints, it is impossible for us to perform deduplication directly on RW-Filtered. Instead, we split CommonCrawl into 100 parts, where each part contains a hundredth of each dump, and perform deduplication on individual parts. Most of the larger duplicate clusters (e.g., licences, common spams) will be shared across parts, and effectively removed. However, we found that CommonCrawl dumps had significant overlap, with URLs being revisited across dumps despite no change in content. Accordingly, we keep a list of the URLs of all samples we have kept from each part, and remove them from subsequent parts being processed.

\section{Experiments}

We now validate that RefinedWeb can be used to train powerful models, matching the zero-shot performance obtained with curated corpora and state-of-the-art language models. We first discuss our evaluation and pretraining setup, and models with which we compare. We perform experiments at small scale to internally compare with other popular datasets, and ablate the three main stages of RefinedWeb (raw, filtered, final). Then, we scale to 1B and 7B models trained on 350GT to compare with state-of-the-art models. Finally, we apply the MDR pipeline to existing pretraining datasets, and show that it can potentially deliver further improvements. 

\subsection{Setting}
\label{sec:exp_setting}

\textbf{Evaluation.} At variance with previous works studying pretraining datasets~\cite{gopher, lee2022deduplicating}, we focus our evaluation on zero-shot generalization across many tasks rather than measuring validation loss. Perplexity alone can be at odds with end-task performance \cite{tay2021scale}, and modern works on LLMs predominantly report zero-shot performance \cite{brown2020language, gopher, chowdhery2022palm}. Furthermore, zero-shot generalization is the ``natural'' setting for autoregressive decoder-only models, in which they perform best \cite{Wang2022WhatLM}. Our evaluation setup is inspired by the one used by the architecture and scaling group of Big Science \cite{scao2022language}.

We base our evaluation on the popular Eleuther AI evaluation harness~\cite{gao2021eval}, allowing us to evaluate across a wide range of tasks in the zero-shot setting. We identified aggregates of tasks allowing us to: (1) obtain signal (i.e., non zero zero-shot performance) at small scale for ablations; (2) compare with results reported by other models. We outline these four aggregates \texttt{small} (for ablations), and \texttt{core}, \texttt{main}, \texttt{ext} (for comparisons) in \cref{tab:task_aggregates}.

Comparisons across models trained and evaluated in different settings are difficult to untangle, as many externalities may influence the 1 
987results (e.g., numerical precision of training vs inference, prompts used). We distinguish three levels of comparisons: (1) internal 
comparisons, with models trained and evaluated within our codebase, for which only the pretraining datasets differ; (2) benchmark-level comparisons, with models trained with a different codebase but evaluated with the Eleuther AI harness, taking results from~\citet{scao2022language, black2022gpt, alephalpha, dey2023cerebras}, thereafter flagged with a $*$; (3) external comparisons  with \citet{brown2020language, chowdhery2022palm}, thereafter flagged with a $\dagger$. For further details on evaluation, see \cref{sec:aggregates}. 

\begin{table*}[t]
    \centering
    \caption{\textbf{Curation is not a silver bullet for zero-shot generalization: small-scale models trained on \textcolor{refinedweb}{\CIRCLE \textsc{RefinedWeb}} outperform models trained on web data (C4, OSCAR), and on curated corpora (\textcolor{pile}{$\blacktriangledown$ The Pile}).} Average accuracy in zero-shot on the \texttt{small-agg} aggregate. All models trained with identical architectures and pretraining hyperparameters. We find that OSCAR-22.01 underperforms other datasets signficantly, perhaps because deduplication is only optional. C4 is a strong baseline, with OSCAR-21.09 lagging slightly behind, but we find that RefinedWeb outperforms both web datasets and the most popular curated dataset, The Pile. Both filtering and deduplication contribute significantly to improving zero-shot performance.}
    \vspace{0.1in}
    \label{tab:small_scale_eai_bs}
    \begin{tabular}{cccccccc}
    \toprule
    & \multicolumn{3}{l}{\textsc{\textbf{Massive web datasets}}} & \multicolumn{1}{l}{\textsc{\textbf{Curated}}} & \multicolumn{3}{l}{\textsc{\textbf{Ours}}} \\ \midrule
       & OSCAR-21.09 & OSCAR-22.01 & C4 & \textcolor{pile}{$\blacktriangledown$ The Pile} & \textcolor{rw_raw}{RW-Raw} & \textcolor{rw_filtered}{RW-Filtered} & \textbf{\textcolor{refinedweb}{\CIRCLE \textsc{RefinedWeb}}}  \\\midrule
        \textbf{1B@27GT} & 55.0\% & 52.7\% & 55.7\% & 53.4\% & 52.7\% & 54.3\% & \textbf{56.2\%} \\
        \textbf{3B@60GT} & 59.1\% & 55.9\% & 59.6\% & 57.9\% & 57.4\% & 58.2\% & \textbf{59.8\%} \\
       \bottomrule
    \end{tabular}
\end{table*}

\textbf{Models.} We train 1B, 3B, and 7B parameters autoregressive decoder-only models, based on configurations and hyperparameters similar to GPT-3 \cite{brown2020language}, diverging mostly on our use of ALiBi \cite{press2021train}. We use FlashAttention \cite{daoflashattention} in a custom codebase. We train internal models on both The Pile and RefinedWeb to control for deviations caused by our pretraining setup--we found The Pile models to perform in-line with others. For small-scale and ablation studies (first half of \cref{sec:web_outperf_curated}; \cref{sec:dedup_curated}), we train models to optimality according to the scaling laws of \citet{hoffmann2022training}:~on 27B and 60B tokens respectively for our 1B and 3B parameters models. For the main experiments demonstrating our approach (Falcon-RW models in \cref{sec:web_outperf_curated}), we train the models to 350GT, in line with popular public models \cite{brown2020language, gpt-j, scao2022bloom}. Note that we do not compare against the recently introduced LLaMA models~\cite{touvron2023llama}, as the smallest of them is trained on x2.5 more compute than our largest model, preventing a meaningful comparison from being made dataset-wise. For a more in-depth overview of the models and pretraining datasets with which we compare, see \cref{sec:other}.

\begin{figure*}[b]
\centering
\subfigure[]{
\includegraphics[width=.45\textwidth]{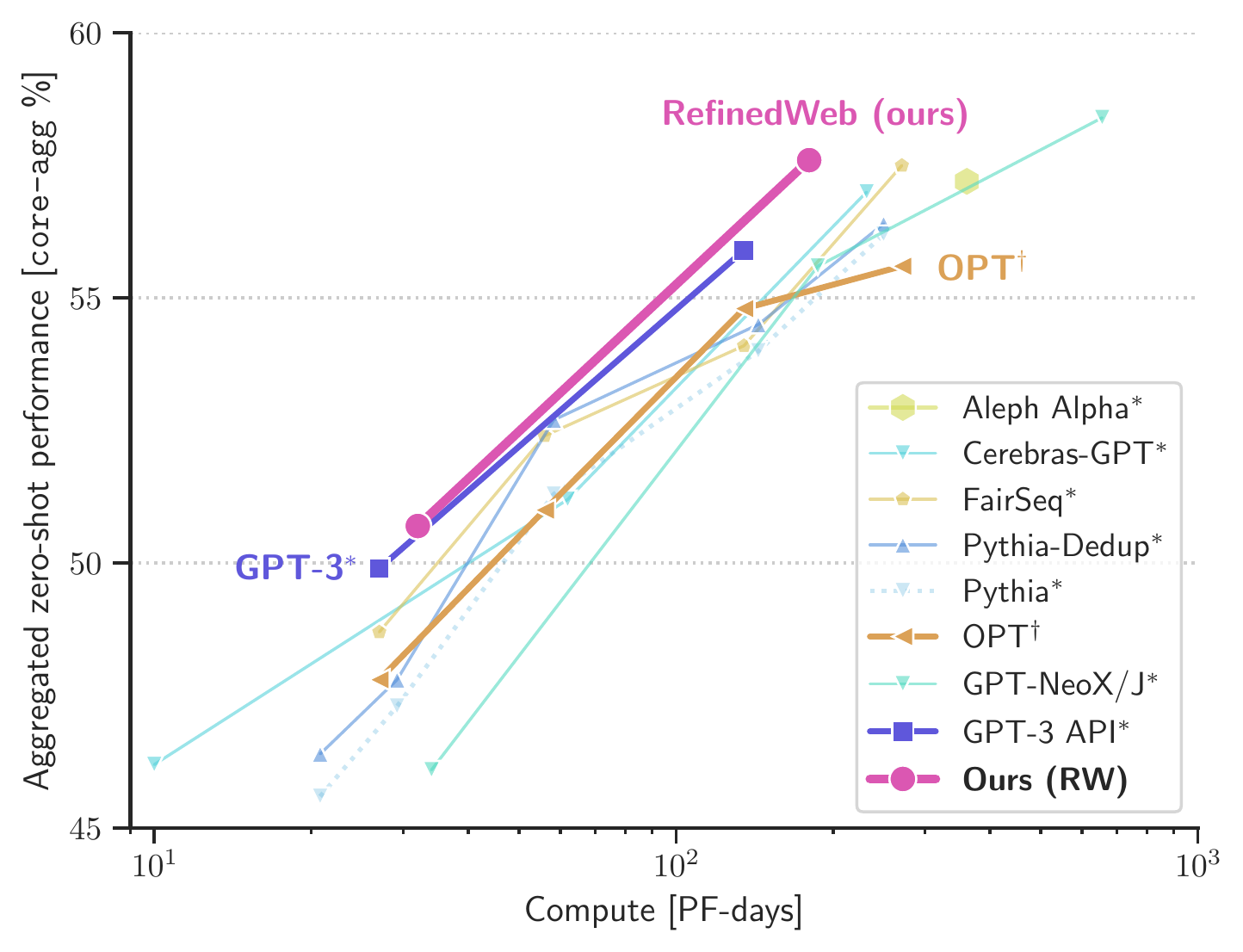}
}
\subfigure[]{
\includegraphics[width=.45\textwidth]{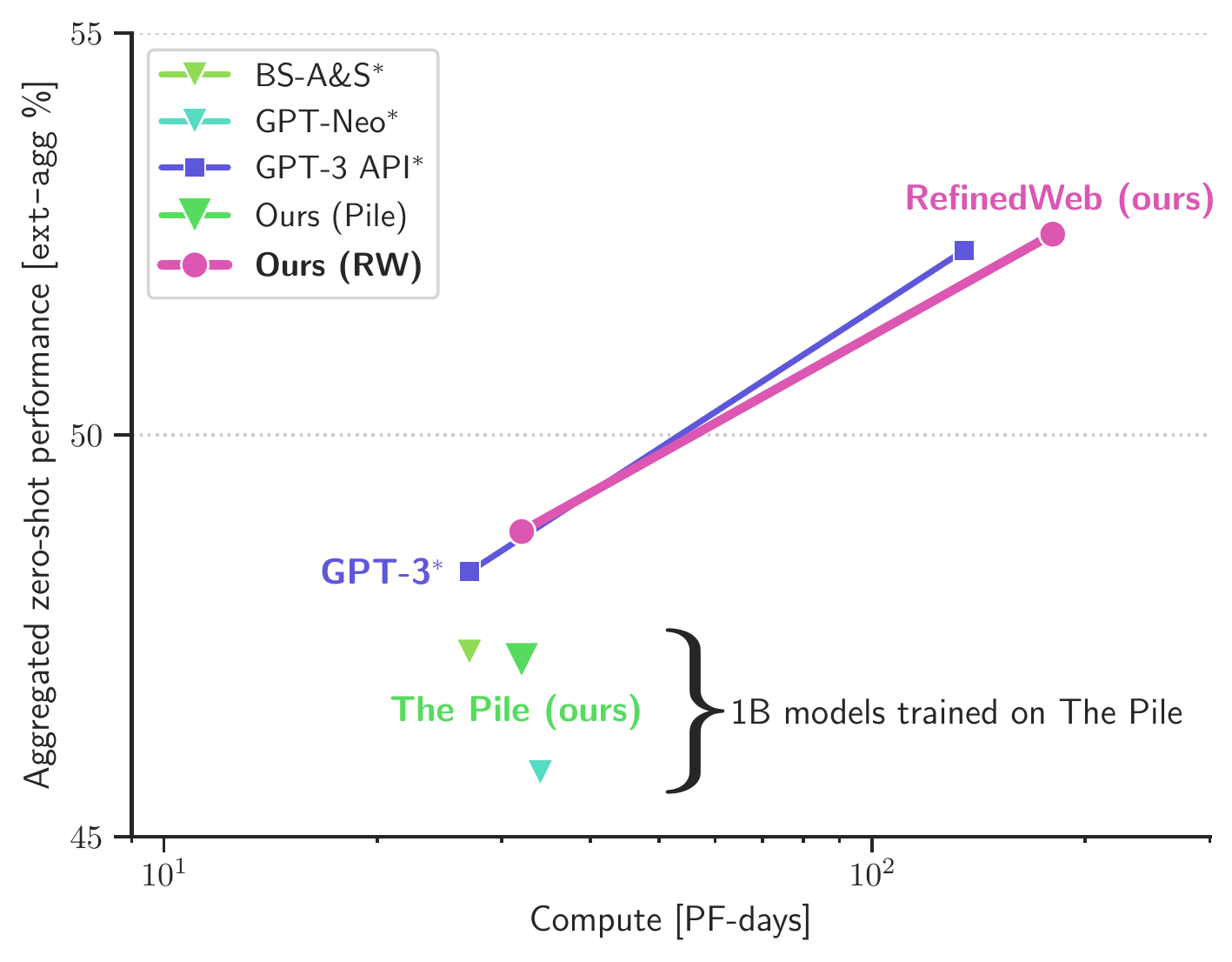}
}
\caption{\textbf{Models trained on \textcolor{refinedweb}{\CIRCLE \textsc{RefinedWeb}} alone outperform models trained on curated corpora.} Zero-shot performance averaged on our~\texttt{core-agg} (left) and \texttt{ext-agg} (right) task aggregates (see \cref{sec:exp_setting} for details, and \cref{fig:main_lead} for results on \texttt{main-agg}). Existing open models fail to match the performance of the original GPT-3 series (left); however, models trained on RefinedWeb significantly outperform models trained on \textcolor{pile}{$\blacktriangledown$ The Pile}: including our direct comparison model (right), ruling out our pretraining setup as the main source of increased performance. In fact, our RefinedWeb models even match the performance of the \textcolor{openai}{$\blacksquare$ GPT-3} models.}
\label{fig:main_ext_zero_shot}
\end{figure*}

\subsection{Can web data alone outperform curated corpora?}
\label{sec:web_outperf_curated}

We endeavour to demonstrate that web data alone can result in models outperforming other models trained on curated corpora. To do so, we first perform a small-scale study with 1B and 3B parameters models trained to optimality~(27GT and 60GT) on popular web and curated datasets. Then, we scale up to 1B and 7B models trained on 350GT, and compare zero-shot generalization to state-of-the-art models.

\paragraph{Small-scale study.} We first consider popular public web datasets (OSCAR-2019 \cite{OrtizSuarezSagotRomary2019}, OSCAR-2022 \cite{AbadjiOrtizSuarezRomaryetal.2021}, C4 \cite{2020t5}), The Pile \cite{gao2020pile} as the most popular publicly available curated dataset, and variations of RefinedWeb (RW-Raw, RW-Filtered, and RW as described in \cref{sec:mdr}). For this first study, all models are trained with the same architecture and the same internal codebase; they are also all evaluated within the same framework--only pretraining datasets differ.

Results averaged on the \texttt{small-=+
} aggregate of 6 tasks are presented in \cref{tab:small_scale_eai_bs}. We observe relatively strong performance of all web datasets compared to The Pile, showcasing that curation is not a silver bullet for performant language models. We find C4 to be a strong pretraining dataset, in line with the findings of \citet{scao2022language}--however, The Pile comparatively underperforms more in our benchmarks. The relatively disappointing results on OSCAR-22.01 may be due to the main version of the dataset being distributed without deduplication. Regarding RefinedWeb, both filtering and deduplication significantly improve performance. 

\paragraph{Full-scale models.} We now validate these results with comparisons with state-of-the-art models. We scale our previous experiments by training 1B and 7B models on 350GT; we also train a 1B model on 350GT on The Pile, as a control for the influence of our pretraining setup. We compare with the following models: the GPT-3 series \cite{brown2020language}, the FairSeq series \cite{artetxe2021efficient}, the GPT-Neo(X)/J models \cite{gpt-neo, gpt-j, black2022gpt}, the OPT series \cite{zhang2022opt}, the BigScience Architecture and Scaling Pile model \cite{scao2022language}, PaLM-8B \cite{chowdhery2022palm},  Aleph Alpha Luminous 13B \cite{alephalpha}, the Pythia series \cite{biderman2023pythia}, and the Cerebras-GPT series \cite{dey2023cerebras}. For GPT-3, we distinguish between results obtained through the API (\texttt{babbage} and \texttt{curie}) with the the EleutherAI LM evaluation harness \cite{gao2021eval} (*), and results reported in their paper, with a different evaluation setup ($\dagger$). Note that for PaLM and OPT, results were also obtained with a different evaluation suite ($\dagger$), while for other models they were obtained with the evaluation harness as well (*), allowing for more direct comparisons. 

Results on \texttt{main-agg} are presented in \cref{fig:main_lead}, and in \cref{fig:main_ext_zero_shot} for \texttt{core-agg} and \texttt{ext-agg}. We find that open models consistently underperform models trained on private curated corpora, such as GPT-3--even when using a similar evaluation setup. Conversely, models trained on RefinedWeb are able to match the performance of the GPT-3 series using web data alone, even though common high-quality sources used in The Pile are excluded from RefinedWeb (see \cref{tab:high-quality-blocked} in Appendix). Finally, we note that our internal model trained on The Pile performs in line with the BigScience Architecture and Scaling model; this highlights that our pretraining setup is unlikely to be the main source of increased performance for models trained on RefinedWeb.

\vfill

\begin{mdframed}
\textbf{Finding.} Challenging existing beliefs on data quality and LLMs, models trained on adequately filtered and deduplicated web data \emph{alone} can match the performance of models trained on curated data.
\end{mdframed}

\vfill

\subsection{Do other corpora benefit from MDR?}
\label{sec:dedup_curated}

Ablating the contributions and evaluating the performance of individual components in the MDR pipeline is difficult: for most heuristics, there is no agreed-upon ground truth, and changes may be too insignificant to result in sufficient zero-shot signal after pretraining. In the first half of \cref{sec:web_outperf_curated}, we identified that subsequent stages of RefinedWeb (raw, filtered, final) led to improvements in performance. In this section, we propose to apply independently the filtering and deduplication stages of MDR to popular pretraining datasets, studying whether they generalize widely.

We report results on the \texttt{small-agg} in \cref{tab:other_mdr}. First, we find that improvements from filtering are not systematic. On The Pile, we had to adjust our line length and characters ratio heuristics to avoid expunging books and code. Despite improvements on OSCAR-21.09, C4, and The Pile, our filters worsen performance on OSCAR-22.01; generally, removal rates from filtering do not seem strongly correlated with downstream accuracy. Conversely, deduplication delivers a steady boost across all datasets, and removal rates are better correlated with changes in performance. We find OSCAR-21.09 and C4 to be already well deduplicated, while The Pile and OSCAR-22.01 exhibit 40-60\% duplicates. The base version of OSCAR-22.01 is distributed without deduplication; for The Pile, this is consistent with the findings of \citet{zhang2022opt}. Finally, combining filtering and deduplication results in further improvements; interestingly, although performance is now more uniform across datasets, differences remain, suggesting that flaws in the original text extraction and processing can't be fully compensated for. 

By processing C4 through MDR, we are able to obtain subsets of data which might slightly outperform RefinedWeb; this combines both the stringent filtering of C4 (e.g., strict NSFW word blocklist, 3-sentence span deduplication) with our own filters and deduplication. While such a combination results in rejection rates that would be unacceptable for our target of 3-6 trillions tokens, this represents an interesting perspective for shorter runs, which may be able to extract extremely high-quality subsets from large web datasets.

\vspace{0.1in}

\begin{mdframed}
\textbf{Finding.} While filtering heuristics may require source-dependent tuning, stringent deduplication improves zero-shot performance across datasets consistently.
\end{mdframed}

\begin{table*}[t]
    \centering
    \caption{\textbf{Although improvements from filtering are not systematic across datasets, deduplication brings a steady performance boost across the board.} Zero-shot accuracy averaged on our \texttt{small-agg} aggregate; [+x.x] reports absolute gains compared to base, removal rates reported against base. Due to limitations in our pipeline, we cannot apply the deduplication stage independently for RefinedWeb.}
    \vspace{0.1in}
    \label{tab:other_mdr}
    \begin{tabular}{lccccc}
    \toprule
     & \multicolumn{3}{l}{\textsc{\textbf{Massive web datasets}}} & \multicolumn{1}{c}{\textsc{\textbf{Curated}}} & \multicolumn{1}{c}{\textsc{\textbf{Ours}}} \\ \midrule
       & OSCAR-21.09 & OSCAR-22.01 & C4 & \textcolor{pile}{$\blacktriangledown$ Pile} & \textcolor{refinedweb}{\CIRCLE RefinedWeb}  \\\midrule
        \textbf{Base} & 55.0\% & 52.7\% & \textbf{55.7\%} & 53.4\% & 52.7\% \\
        \textbf{Filtered} & 55.4\% [+.4] & 52.3\% [-.4] & \textbf{56.2\%} [+.5] & 54.2\% [+.8] & 54.3\% [+1.6] \\
        \emph{removal rate} & \emph{-25.0\%} & \emph{-39.8\%} & \emph{-16.4\%} & \emph{-27.1\%} & \emph{-50.8\%} \\
        \textbf{Deduplicated} & 55.6\% [+.6] & 55.6\% [+2.9] & \textbf{55.9\%} [+.2] & 54.5\% [+1.1] & \\
        \emph{removal rate} & \emph{-10.8\%} & \emph{-60.8\%} & \emph{-7.59\%} & \emph{-45.3\%} & \\
        \textbf{Filt.+Dedup.} & 55.5\% [+.5] & 55.4\% [+2.7] & \textbf{56.4\%} [+.7] & 55.2\% [+1.8] & 56.2\% [+3.5]\\
        \emph{removal rate} & \emph{-28.2\%} & \emph{-62.2\%} & \emph{-17.9\%} & \emph{-66.0\%} & \emph{-75.4\%}\\
       \bottomrule
    \end{tabular}
    \vspace{-0.1in}
\end{table*}

\section{Limitations}
\begin{figure}[h]
\centering
\includegraphics[width=0.37\textwidth]{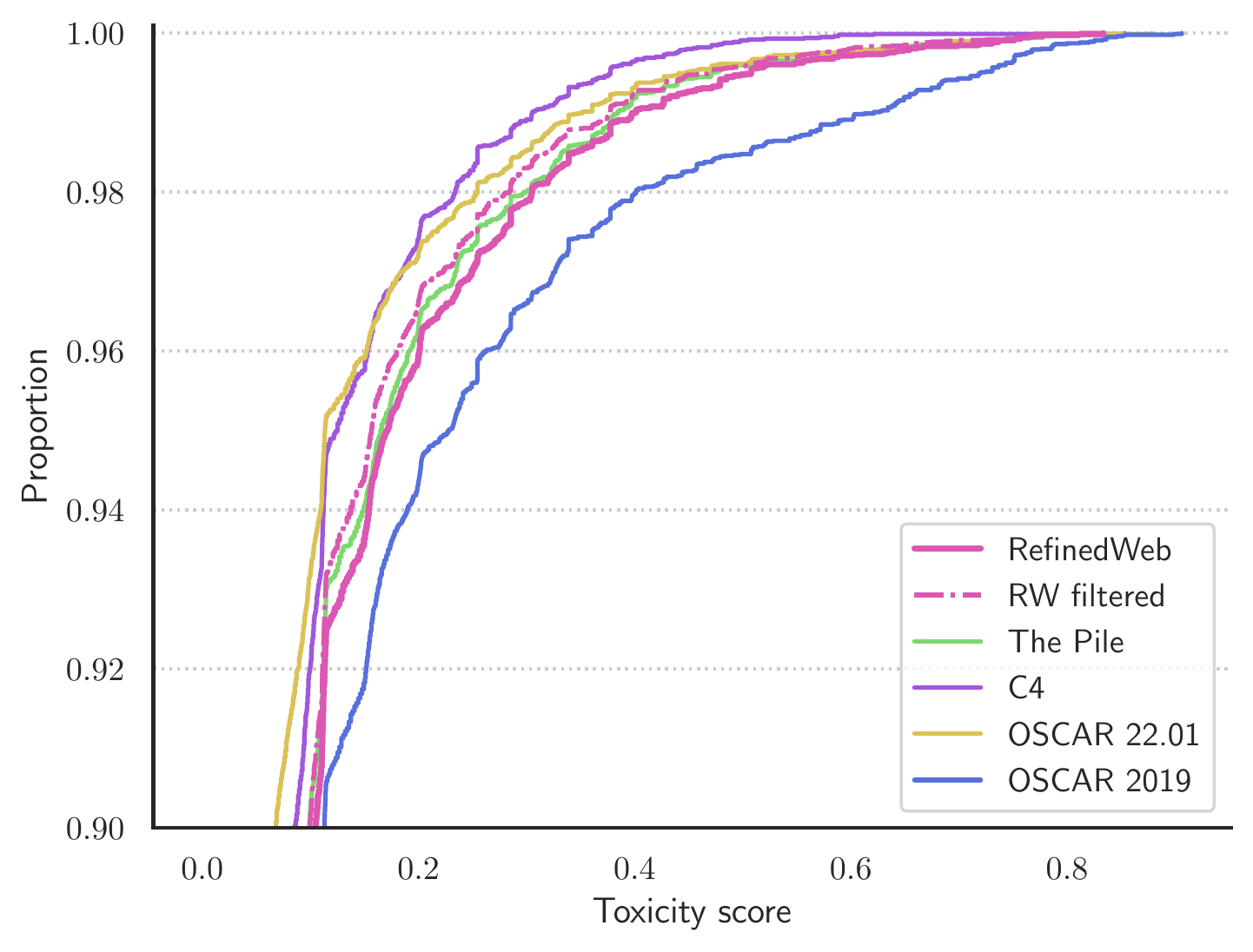}
\caption{\textbf{Toxic content in \textcolor{refinedweb}{RefinedWeb} is distributed similarly to \textcolor{pile}{The Pile.}} Cumulative proportion of documents below a given toxicity score, as evaluated by the Pespective API.}
\label{fig:toxicity_main}
\end{figure}

\paragraph{Biases.} We conduct a basic analysis of the toxicity of RefinedWeb in \cref{fig:toxicity_main}. We find RW to be about as toxic as The Pile, based on the definition of toxicity provided by the Perspective API: "content that is rude or disrespectful". Notably, this definition does not cover issues with social biases or harmfulness. Although it is unlikely that our pipeline introduces further issues on this side than is already documented for popular datasets, we encourage further quantitative work on the public extract of RefinedWeb.

\paragraph{Multiple epochs.} Instead of looking for "unique" tokens to make up a trillion-scale pretraining dataset, one could simply repeat data over multiple epochs. Popular models like OPT and NeoX-20B do this for up to 2 epochs, and most curated datasets upsample corpora 2-5 times. However,~\citet{hernandez2022scaling} has recently shown that models with 100B+ parameters may be sensitive to even just a few epochs. Orthogonal to our work lies a line of research exploring tradeoffs in the data-constrained regime: can deduplication help sustain more epochs? Are multiple epochs on higher quality data better than a one epoch on lower quality data? See \cref{sec:dedup_epochs} for a more in-depth discussion.

\paragraph{Other results on deduplication.} \citet{biderman2023pythia} found a limited impact on zero-shot performance from deduplicating The Pile; we discuss further in \cref{sec:other_models}, but encourage further deduplication research on curated corpora, and studying deduplication in the data-constrained regime, where multiple epochs have to be performed to compensate for the reduction in tokens incurred by deduplication.

\vspace{-0.1in}
\section{Conclusion}
As LLMs are widely adopted, models trained past the recommendations of scaling laws are bound to become increasingly common to amortize inference costs~\cite{touvron2023llama}. This will further drive the need for pretraining datasets with trillions of tokens, an order of magnitude beyond publicly available corpora. We have demonstrated that stringent filtering and deduplication could result in a five trillion tokens web only dataset suitable to produce models competitive with the state-of-the-art, even outperforming LLMs trained on curated corpora. We publicly release a 600GT extract of RefinedWeb, and note that RefinedWeb has already been used to train state-of-the-art language models, such as Falcon-40B \cite{falcon40b}.

\bibliography{example_paper}
\bibliographystyle{icml2023}

\newpage
\appendix
\onecolumn

\newpage

\section{RefinedWeb Datasheet}
\begin{longtable}{p{6cm}|p{10cm}}
    \toprule
    \multicolumn{2}{c}{\textsc{\textbf{Motivation}}} \\
    \midrule
    \textbf{For what purpose was the dataset created?} & RefinedWeb was created to serve as a large-scale dataset for the pretraining of large language models. It may be used on its own, or augmented with curated sources (e.g., Wikipedia, StackOverflow). \\ \midrule
    \textbf{Who created the dataset and on behalf of which entity?} & The dataset was created by the Technology Innovation Institute.\\ \midrule
    \textbf{Who funded the creation of the dataset?} & The creation of the dataset was privately funded by the Technology Innovation Institute. \\ \midrule
    \textbf{Any other comment?} & RefinedWeb is built on-top of CommonCrawl, using the Macrodata Refinement Pipeline, which combines content extraction, filtering heuristics, and deduplication. In designing RefinedWeb, we abided to the following philosophy: (1) \textbf{Scale first.} We intend MDR to produce datasets to be used to train 40-200B parameters models, thus requiring trillions of tokens \cite{hoffmann2022training}. For English-only RefinedWeb, we target a size of 3-6 trillion tokens. Specifically, we eschew any labour intensive human curation process, and focus on CommonCrawl instead of disparate single-domain sources. (2) \textbf{Strict deduplication.} Inspired by the work of \citet{lee2022deduplicating}, which demonstrated the value of deduplication for large language models, we implement a rigorous deduplication pipeline. We combine both exact and fuzzy deduplication, and use strict settings leading to removal rates far higher than others have reported. (3) \textbf{Neutral filtering.} To avoid introducing further undesirable biases into the model \cite{dodge2021documenting, welbl2021challenges}, we avoid using ML-based filtering outside of language identification. We stick to simple rules and heuristics, and use only URL filtering for adult content. \\ \midrule
    \multicolumn{2}{c}{\textsc{\textbf{Composition}}} \\ \midrule
    \textbf{What do the instances that comprise the dataset represent?} & Instances are text-only documents, corresponding to single web pages. \\ \midrule
    \textbf{How many instances are there in total?} & RefinedWeb contains $\sim$10 billion documents, or around 5 trillion tokens. The public version is a subset representing a tenth of the full version. \\ \midrule
    \textbf{Does the dataset contain all possible instances or is it a sample (not necessarily random) of instances from a larger set?} & RefinedWeb is built using all CommonCrawl dumps until the 2023-06 one; it could be updated with additional dumps as they are released. The public release of RefinedWeb is a 600GT random extract of the 5,000GT of the full dataset. For all experiments, we randomly sampled from the public extract, or earlier development versions of it. \\ \midrule
    \textbf{What data does each instance consist of?} & Each instance is a text-only document, with metadata about its origin in CommonCrawl and source page URL. We also distribute a multimodal version of RefinedWeb, containing interlaced links to images. \\ \midrule
    \textbf{Is there a label or target associated with each instance?} & No. \\ \midrule
    \textbf{Is any information missing from individual instances?} & No. \\ \midrule
    \textbf{Are relationships between individual instances made explicit?} & No. \\ \midrule
    \textbf{Are there recommended data splits?} & No. \\ \midrule
    \textbf{Are there any errors, sources of noise, or redundancies in the dataset?} & Despite our best efforts to filter content that does not qualify as natural language, and to deduplicate documents, our pipeline may let through documents that may be considered as errors or redundant. \\ \midrule
    \textbf{Is the dataset self-contained, or does it link to or otherwise rely on external resources?} & The base version of the dataset is self-contained, but the multimodal version is interlaced with links to images--these are not distributed as part of the dataset, and constitute an external source. \\ \midrule
    \textbf{Does the dataset contain data that might be considered confidential?} & All documents in RefinedWeb have been publicly available online. \\ \midrule
    \textbf{Does the dataset contain data that, if viewed directly, might be offensive, insulting, threatening, or might otherwise cause anxiety?} & Yes, as this type of data is prevalent on the internet, it is likely our dataset contains such content. Notably, we estimate the prevalence of toxic content in the dataset to be similar to The Pile (\cref{fig:toxicity_main}). \\ \midrule
    \multicolumn{2}{c}{\textsc{\textbf{Collection}}} \\ \midrule
    \textbf{How was the data associated with each instance acquired?} & We downloaded with \texttt{warcio} publicly available .WET files from the CommonCrawl foundation. \\ \midrule
    \textbf{What mechanisms or procedures were used to collect the data?} & We refer to the CommonCrawl website (\url{commoncrawl.org}) for details on how they collect data. \\ \midrule
    \textbf{If the dataset is a sample from a larger set, what was the sampling strategy?} & Whenever we use subsets, we randomly sample from the original data. \\ \midrule
    \textbf{Who was involved in the data collection process and how were they compensated?} & The original data collection was performed by CommonCrawl; authors from this paper were involved in retrieving it and preparing it. \\ \midrule
    \textbf{Over what timeframe was the data collected?} & We use all CommonCrawl dumps from 2008 to January/February 2023. \\ \midrule
    \textbf{Were any ethical review processes conducted?} & No. \\ \midrule
    \multicolumn{2}{c}{\textsc{\textbf{Preprocessing}}} \\ \midrule
    \textbf{Was any preprocessing/cleaning/labeling of the data done?} & Yes, we applied extensive preprocessing and cleaning of the data. We first filter URLs to remove adult content using a blocklist and a score system (\cref{sec:url_details}), we then use \texttt{trafilatura} \cite{barbaresi-2021-trafilatura} to extract content from pages, and perform language identification with the \texttt{fastText} classifier from CCNet \cite{wenzek2020ccnet}. After this first preprocessing stage, we filter data using heuristics from MassiveWeb~\cite{gopher} and our own line-wise corrections (\cref{sec:line_details}). Finally, we run extensive deduplication, removing URLs revisited across dumps (\cref{sec:dedup_mdr}) and performing subsequently fuzzy and exact substring deduplication, with each stage drawing from \citet{lee2022deduplicating}.  See \cref{sec:mdr} for further details and \cref{tab:mdr_pipeline} for an outline.   \\ \midrule
    \textbf{Was the “raw” data saved in addition to the preprocessed/cleaned/labeled data?} & During development, we saved intermediary outputs from our pipeline for investigations and for ablations--intermediary outputs exist for about 5\% of RefinedWeb. We did not keep intermediary outputs for the final production version of the dataset due to storage and resource constraints. \\ \midrule
    \textbf{ Is the software that was used to preprocess/clean/label the data available?} & No. \\ \midrule
    \multicolumn{2}{c}{\textsc{\textbf{Uses}}} \\ \midrule
    \textbf{Has the dataset been used for any tasks already?} & Yes, this data has been used to develop large language models: both for scientific experiments (e.g., this paper) and production use. \\ \midrule
    \textbf{Is there a repository that links to any or all papers or systems that use the dataset?} & No. \\ \midrule
    \textbf{What (other) tasks could the dataset be used for?} & RefinedWeb was built as a large-scale corpora representative of the web, and as such may see many downstream uses which are difficult to predict. \\ \midrule
    \textbf{Is there anything about the composition of the dataset or the way it was collected and preprocessed/cleaned/labeled that might impact future uses?} & For the public extract of RefinedWeb, we chose to only draw from the English version of the dataset, preventing multilingual applications. \\ \midrule
    \textbf{Are there tasks for which the dataset should not be used?} & Any tasks which may considered irresponsible or harmful. \\ \midrule
    \multicolumn{2}{c}{\textsc{\textbf{Distribution}}} \\ \midrule
    \textbf{Will the dataset be distributed to third parties outside of the entity on behalf of which the dataset was created?} & Yes, we make a 600GT extract publicly available for NLP practitioners. We currently don't plan to share the full version of the dataset. \\ \midrule
    \textbf{How will the dataset will be distributed?} & The dataset will be made available through the HuggingFace Hub. \\ \midrule
    \textbf{When will the dataset be distributed?} & The dataset is available immediately. \\ \midrule
    \textbf{Will the dataset be distributed under a copyright or other intellectual property (IP) license, and/or under applicable terms of use (ToU)?} & The public extract is made available under an ODC-By 1.0 license; users should also abide to the CommonCrawl ToU: \url{https://commoncrawl.org/terms-of-use/}. \\ \midrule
    \textbf{Have any third parties imposed IP-based or other restrictions on the data associated with the instances?} & Not to our knowledge. \\ \midrule
    \textbf{Do any export controls or other regulatory restrictions apply to the dataset or to individual instances?} & Not to our knowledge. \\ \midrule
    \multicolumn{2}{c}{\textsc{\textbf{Maintenance}}} \\ \midrule
    \textbf{Who will be supporting/hosting/maintaining the dataset?} & The dataset will be hosted on the HuggingFace Hub, we have no plans to further support or maintain it once it is released. \\ \midrule
    \textbf{How can the owner/curator/manager of the dataset be contacted?} & falconllm@tii.ae \\ \midrule
    \textbf{Is there an erratum?} & No. \\ \midrule
    \textbf{Will the dataset be updated?} & No. \\ \midrule
    \textbf{If others want to extend/augment/build on/contribute to the dataset, is there a mechanism for them to do so?} & No. \\ \bottomrule
    \caption{\textbf{Datasheet for RefinedWeb}, following the framework introduced by \citet{gebru2021datasheets}.}
    \label{tab:datasheet}
\end{longtable}

\newpage

\section{Falcon-RW Model Cards}
\begin{longtable}{p{6cm}|p{10cm}}
    \toprule
    \multicolumn{2}{c}{\textsc{\textbf{Model details}}} \\
    \midrule
    \textbf{Person/organization developing the model} & The models were created by the Technology Innovation Institute. \\ \midrule
    \textbf{Model date} & Falcon-RW models were trained in December 2022/January 2023. \\ \midrule
    \textbf{Model type and information about training} & Falcon-RW are autoregressive Transformer models trained with a causal language modeling objective. Architecture based on GPT-3 \cite{brown2020language}, with ALiBi positional encodings \cite{press2021train} and~FlashAttention \cite{daoflashattention}. See \cref{sec:exp_setting} for details. \\ \midrule
    \textbf{Licence} & Apache 2.0: \url{https://www.apache.org/licenses/LICENSE-2.0}. \\ \midrule
    \textbf{Point of contact} & falconllm@tii.ae \\ \midrule
    \multicolumn{2}{c}{\textsc{\textbf{Intended use}}} \\ \midrule
    \textbf{Primary intended uses} & Research on large language models, and the influence of adequately filtered and deduplicated web data on the properties of large language models (fairness, safety, limitations, capabilities, etc.). \\ \midrule
    \textbf{Primary intended users} & NLP researchers. \\ \midrule
    \textbf{Out-of-scope use cases} & Production use without adequate assessment of risks and mitigation; any use cases which may be considered irresponsible or harmful. \\ \midrule
    \multicolumn{2}{c}{\textsc{\textbf{Factors}}} \\ \midrule
    \textbf{Relevant factors} & Falcon-RW models are trained on English data only, and will not generalize appropriately to other languages. Furthermore, as they are trained on a large-scale corpora representative of the web, they will carry the stereotypes and biases commonly encountered online. \\ \midrule
    \textbf{Evaluation factors} & We evaluated the toxicity of the underlying pretraining dataset and found it to be in line with common curated pretraining datasets such as The Pile (see \cref{fig:toxicity_main}). Note that this only accounts for toxicity under the definition of Perspective API: "content that is rude or disrespectful". Notably, this fails to include concerns about social biases or harmfulness. \\ \midrule
    \multicolumn{2}{c}{\textsc{\textbf{Metrics}}} \\ \midrule
    \textbf{Model performance measures} & We focus our evaluation on measuring the zero-shot generalization capabilities of our models across a wide range of tasks, leveraging the Eleuther AI language model evaluation harness \cite{gao2021eval}.  \\ \midrule
    \textbf{Variation approaches} & Due to the costs associated with training Falcon-RW we cannot train the models multiple times and measure variability across training runs. \\ \midrule
    \multicolumn{2}{c}{\textsc{\textbf{Evaluation data}}} \\ \midrule
    \textbf{Datasets} & We evaluate zero-shot accuracy on 18 varied tasks, detailed in \cref{tab:task_aggregates}. \\ \midrule
    \textbf{Motivation} & We selected and aggregated tasks to build comparisons with other models in the literature (see \cref{sec:exp_setting}; \cref{sec:aggregates} for details). \\ \midrule
    \textbf{Preprocessing} & We use the default prompts and setup of \citet{gao2021eval}. \\ \midrule
    \multicolumn{2}{c}{\textsc{\textbf{Training data}}} \\ \midrule
    \multicolumn{2}{c}{\textbf{See the dedicated datasheet in \cref{tab:datasheet}.}} \\
    \bottomrule
    \caption{\textbf{Model card for Falcon-RW}, following the framework introduced by \citet{mitchell2019model}.}
    \label{tab:model_card}
\end{longtable}

\section{Dataset analysis}

\begin{figure}[h]
\centering     
\subfigure[Document Lengths]{\label{fig:document_lengths}\includegraphics[width=0.4\textwidth]{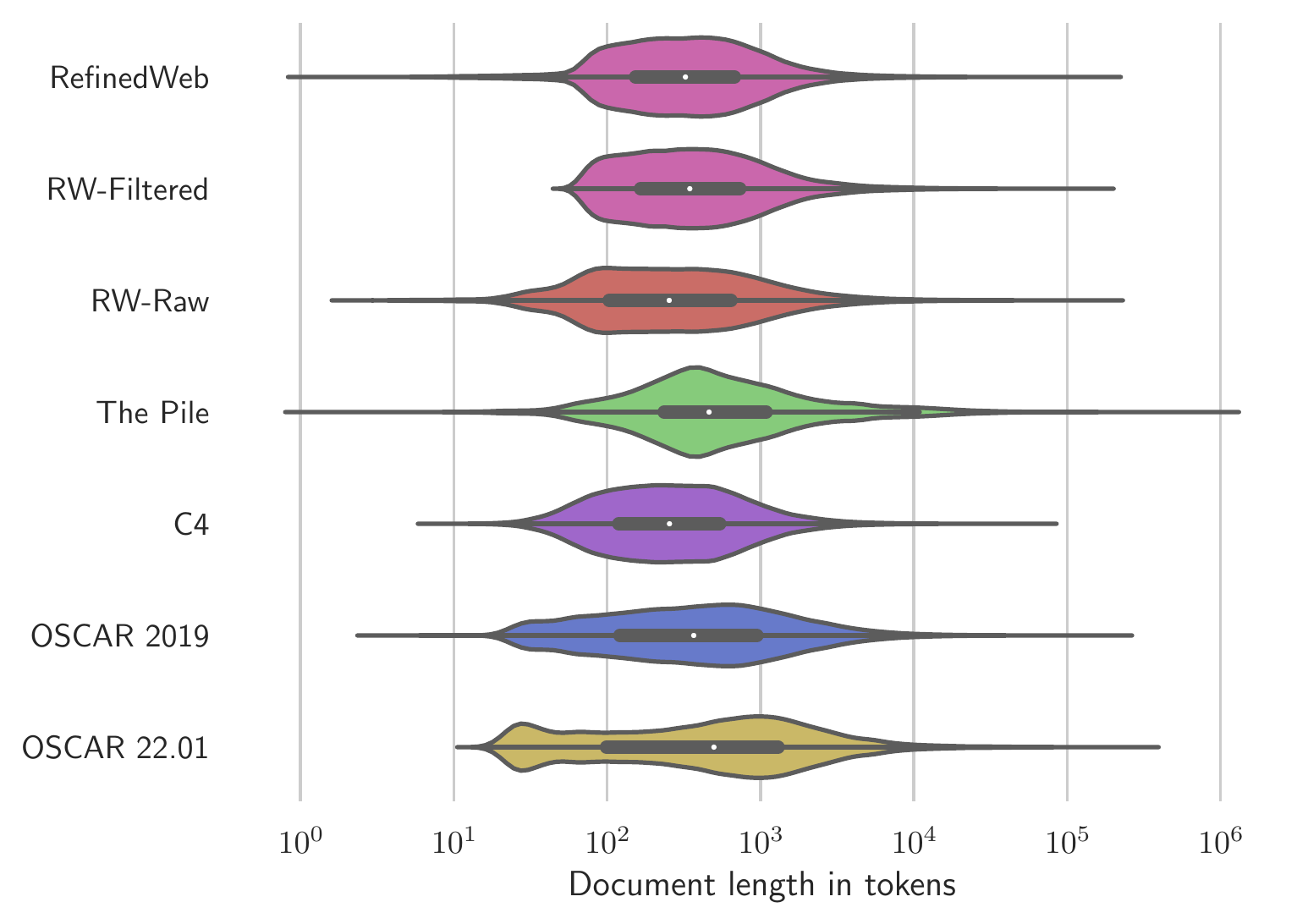}}
\subfigure[Top domains]{\label{fig:refinedweb_domain_breakdown}\includegraphics[width=0.5\textwidth]{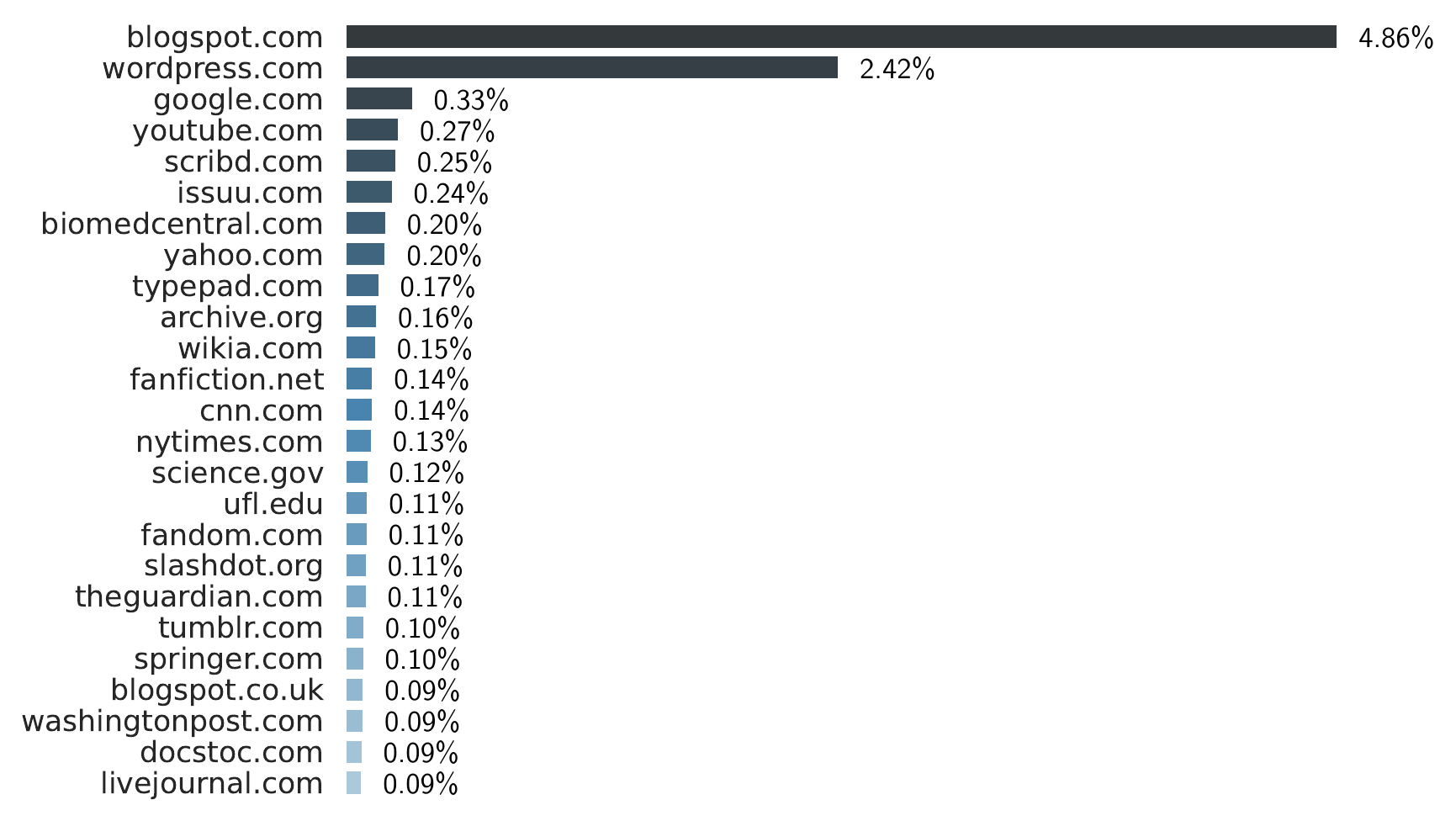}}
\caption{\textbf{Make-up of \textcolor{refinedweb}{RefinedWeb} in document lengths (left) and top domains (right).} (a) We find the OSCAR datasets and \textcolor{rw_raw}{RW-Raw} to have similar document length distributions; following filtering, most of the short documents are discarded from \textcolor{rw_filtered}{RW-Filtered}. As deduplication removes spans, it reintroduces shorter documents to \textcolor{refinedweb}{RefinedWeb}. We note the make-up of C4 and RefinedWeb to be relatively similar, with a longer tail of short documents for RefinedWeb. Finally, \textcolor{pile}{The Pile} exhibit a unique make-up, with a long tail of both long (books, etc.) and short documents.  (b) Top domains in RefinedWeb span from popular content platforms (Blogspot, WordPress, Tumblr, etc.), to news websites (CNN, New York Times, etc.), and include also technical content such as BioMed Central or Springer.}
\end{figure}

The large-scale and diverse nature of web corpora make them difficult to document and analyse as a whole; we provide some key metrics in the section, focusing on document lengths in \cref{fig:document_lengths}, and a breakdown of the top domain names in \cref{fig:refinedweb_domain_breakdown}. We also refer to the analysis of the distribution of toxic content presented in \cref{fig:toxicity_main}.

\section{Multilingual RefinedWeb}
\label{sec:multilingual}

\begin{figure}[h]
\centering
\includegraphics[width=0.9\linewidth]{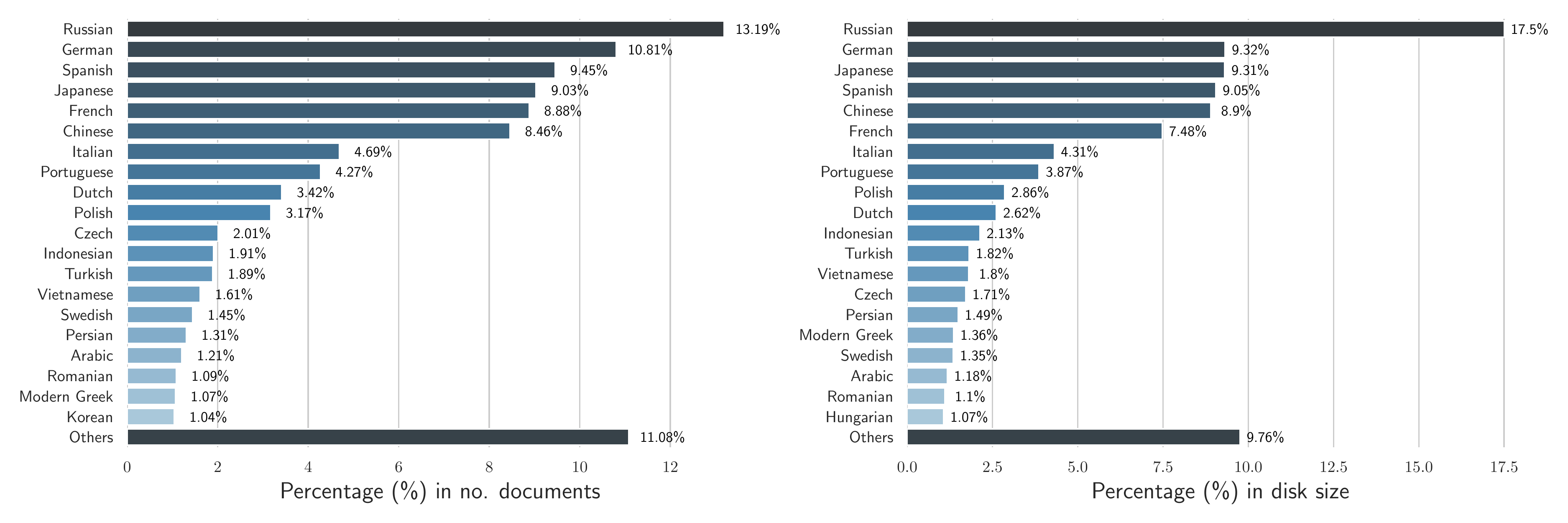}
\caption{\textbf{Top 20 languages (excluding English) from processed CommonCrawl based on number of documents and disk size.}}
\label{fig-ml-lang-dist}
\end{figure}

\paragraph{Multilingual data.} Using the language identification filter, we classify processed CommonCrawl data into 176 languages. Figure~\ref{fig-ml-lang-dist} shows the top 20 languages present in the data \textit{excluding English}, based on their relative contribution in descending order. 58.20\% of all documents in the processed CommonCrawl data were identified as English. We find the distribution of languages in CommonCrawl to only be partially aligned with the worldwide distribution of language speakers \cite{ethnologue}: Russian is over-represented (2nd in CC but only 8th worldwide), Mandarin Chinese is under-represented (6-7th in CC but 2nd worldwide), and Hindi does not show-up in the top 20 despite being the 3rd most spoken. 

\vspace{0.1in}

\paragraph{Processing multilingual data.}  The MDR pipeline can be used to process all languages: features such as text extraction are language-agnostic, whereas specific filters such as line-wise corrections need to typically be tuned for each individual language. We also found tuning deduplication parameters for individual languages to be beneficial.

\section{Additional results}

In this section, we present additional results obtained during the development of the Macrodata Refinement pipeline. For \cref{sec:ablation_dedup} and \cref{sec:dedup_epochs}, these were obtained using earlier development versions of the dataset, so results are not directly comparable with the main text. For \cref{sec:lm_eval}, this is based on the Falcon-RW models.

\subsection{Small-scale ablations on deduplication approaches}
\label{sec:ablation_dedup}

We present results in \cref{tab:dedup_ablations}--the setup is similar to our earlier ablations, training 1B models for 30GT. We observe that:
\begin{itemize}
    \item \textbf{MinHash alone is insufficient}, as it doesn't match the zero-shot performance of exact deduplication. Conversely, combining it with exact deduplication doesn't improve performance further. 
    \item \textbf{Masking spanned duplicates degrades performance}, systematically underperforming other approaches. Dropping and cutting spans perform similarly, although it's likely that dropping documents slightly outperforms cutting.
\end{itemize}

Finally, we chose to apply MinHash before exact deduplication, as it is easier to scale: approximate deduplication acts as a pruning phase, enabling us to scale deduplication further. Finally, we choose the common option of cutting spans, as dropping resulted in even more stringent rejection rates which would have compromised our ability to collect 5 trillion tokens.

\begin{table}[h]
\centering
    \caption{\textbf{MinHash alone is insufficient to match the performance of exact substring deduplication, and combining the two does not significantly improve performance. Of all of the exact substring approaches, masking duplicated spans underperform, but all others exhibit similar performance.} $\checkmark$ Minhash + Exact substring-Cut corresponds to our final deduplication setup. Perplexity in bits-per-bytes on The Pile (\texttt{pile-bpb}, lower is better), zero-shot performance aggregated over LAMBADA, PIQA, and HellaSwag (\texttt{agg-dev}). Best results in \textbf{bold}, best results with minhash in \underline{underline}, table sorted by increasing \texttt{agg-dev-1}.}
    \label{tab:dedup_ablations}
\vspace{0.1in}
\centerline{\begin{tabular}{cccc}
\toprule
\textbf{Minhash} & \textbf{Exact substring} & \texttt{pile-bpb} $\downarrow$ & \texttt{agg-dev-1} $\uparrow$ \\ \midrule
\multicolumn{2}{c}{\textcolor{rw_filtered}{RefinedWeb-Filtered}} & 1.11 & 43.51 \\ \midrule
& Mask & 1.08 & 45.84 \\
\checkmark & Mask & 1.07 & 46.28 \\
\checkmark &  & 1.07 & 46.57 \\
\textcolor{refinedweb}{\checkmark} & \textcolor{refinedweb}{Cut} & \underline{\textbf{1.05}} & 47.11 \\
& Cut & 1.06 & 47.24 \\
\checkmark & Drop partial & \underline{\textbf{1.05}} & 47.25 \\
& Drop any & 1.07 & 47.77 \\
\checkmark & Drop any & 1.07 & \underline{47.86} \\
& Drop partial & 1.06 & \textbf{47.97} \\ \midrule
\multicolumn{2}{c}{\textcolor{pile}{Pile}} & 0.88 & 43.70 \\ \bottomrule
\end{tabular}}
\end{table}

\subsection{Language modeling evaluation}
\label{sec:lm_eval}

Along with our aggregates, we also evaluated perplexity on Wikitext (\cref{tab:lm_supplementary}). We found that models trained on RefinedWeb achieve performance close to that of models trained on The Pile. Importantly, we note that RefinedWeb does not contain any content from Wikipedia -- it is explicitly filtered out at the URL level. We believe this accounts for most of the difference in perplexity, as RW models may not be familiar with the idiosyncrasies of Wikitext (e.g., layout of an article, etc.)

\begin{table}[h]
\centering
    \caption{\textbf{Models trained on \textcolor{refinedweb}{RefinedWeb} achieve performance close to models trained on \textcolor{pile}{The Pile} on Wikitext, despite not having seen any content from Wikipedia.} Perplexity in bits-per-bytes on Wikitext (\texttt{wiki-bpb}, lower is better.)}
    \label{tab:lm_supplementary}
\vspace{0.1in}
\centerline{\begin{tabular}{lccc}
\toprule
\textbf{Model size} & \textbf{1B} & & \textbf{7B} \\
\textbf{Dataset} & \textcolor{pile}{\textbf{The Pile}} & \textcolor{refinedweb}{\textbf{RW}} & \textcolor{refinedweb}{\textbf{RW}} \\\midrule
\texttt{wiki-bpb} $\downarrow$ & 0.64 & 0.66 & 0.60 \\ \bottomrule
\end{tabular}}
\end{table}

\vspace{-0.1in}
\subsection{Does deduplication help with multiple epochs?}
\label{sec:dedup_epochs}

Earlier in this work, we outlined that to scale pretraining data, practitioners had two choices: (1) improve data collection, which is the avenue we chose to pursue; (2) train models on multiple epochs of the same data. Due to current uncertainties in the ability of larger models to sustain multiple epochs without adverse effects \cite{hernandez2022scaling}, we focused on (1). A fairly rational question regarding (2) is whether deduplication may improve the situation, and whether deduplicated data may be able to sustain more epochs without compromising model quality. 

\begin{figure}[h]
\centering     
\subfigure[Degradation compared to 1 epoch]{\label{fig:epochs_degradation}\includegraphics[width=0.45\textwidth]{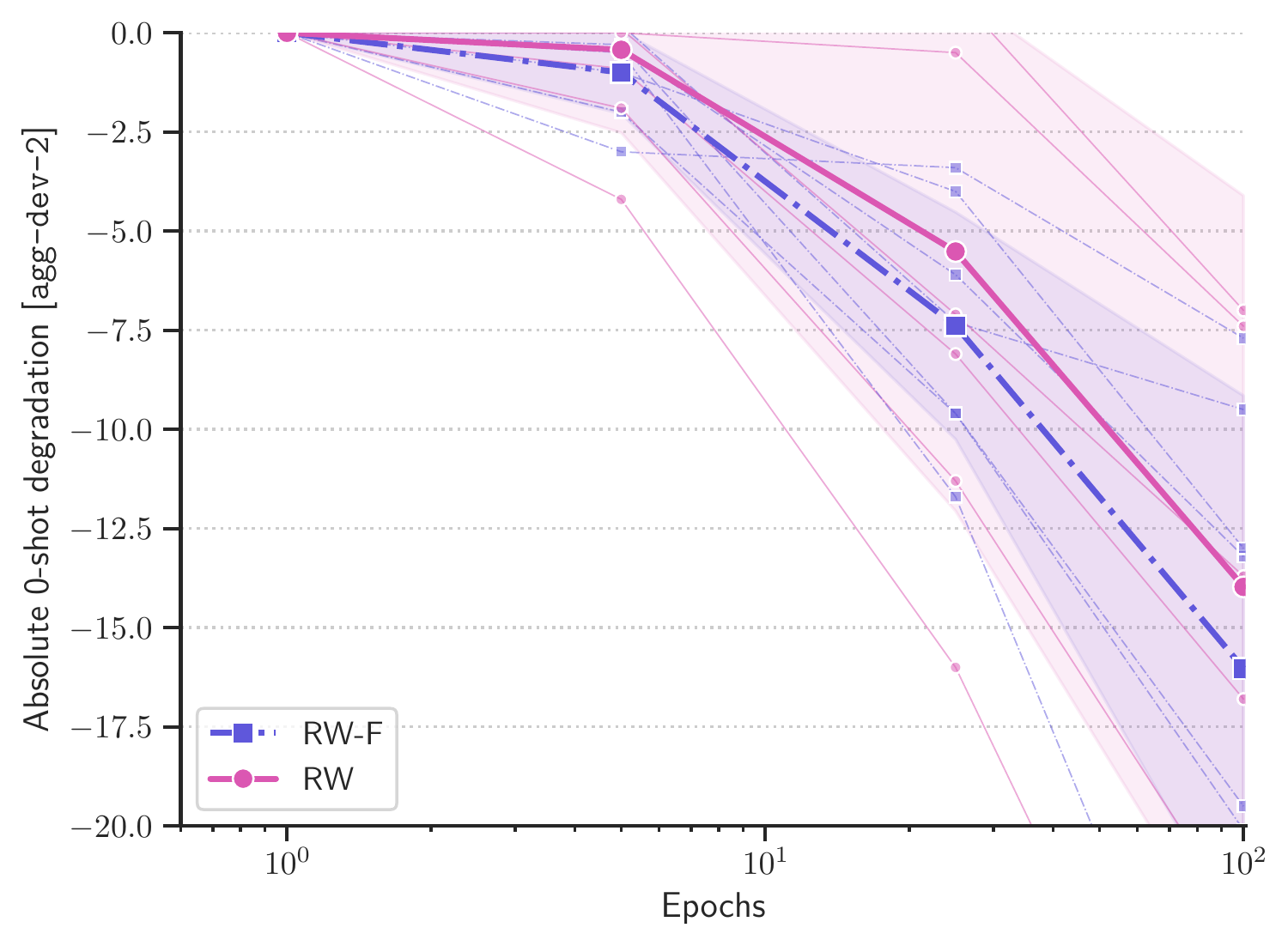}}
\subfigure[Gap between RW and RW-F]{\label{fig:epochs_gap}\includegraphics[width=0.43\textwidth]{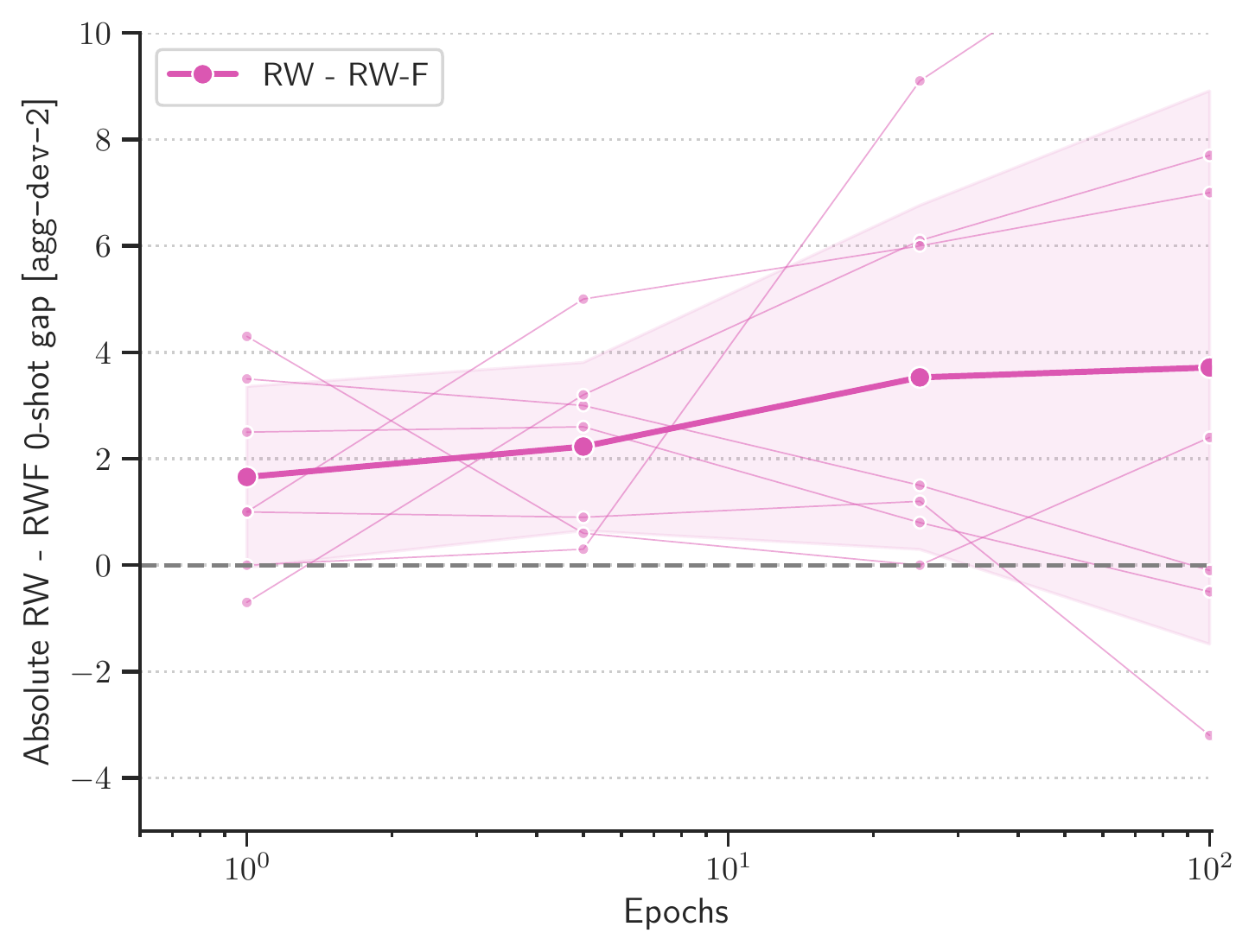}}
\caption{\textbf{Deduplication may reduce the degradation in performance incurred by multiple epochs.} However, our experiments were only performed at small-scale (1B models trained on 30GT), and we see high variability in outcomes across tasks. Zero-shot performance measured on the \texttt{agg-dev-2} aggregate (HellaSwag, PIQA, ARC, BoolQ, COPA, MRPC, SciQ). Individual curves for per-task results and 1-$\sigma$ standard deviation across all tasks in the aggregate in transparent.}
\end{figure}

We train 1B parameters models on 30GT of RW and RW-Filtered. We keep the number of pretraining tokens fixed, but train for 1, 5, 25, and 100 epochs. This is a small-scale, limited set-up, which would have to be improved to obtain definitive results. We plot the degradation in performance compared to a single epoch in \cref{fig:epochs_degradation} and the gap between RW and RW-F in \cref{fig:epochs_gap}. We find that the absolute degradation is less important for RefinedWeb than for RefinedWeb-Filtered; furthermore, the gap widens with increasing number of epochs. However, we observe significant variability across tasks.

\section{Tasks, models, and datasets from the state-of-the-art}
\label{sec:other}

\subsection{Task aggregates}
\label{sec:aggregates}

To evaluate models, we average zero-shot performance over diverse task aggregates Our aggregates are outlined in \cref{tab:task_aggregates}:
\begin{itemize}
    \item \texttt{small}: small-scale ablation studies, taskswith non-zero performance for 1B parameters models trained on 30GT;
    \item \texttt{core}: comparisons with a wide range of models, notably based on the tasks reported in \cite{dey2023cerebras};
    \item \texttt{main}: tasks available in the GPT-3 and PaLM papers \cite{brown2020language, chowdhery2022palm};
    \item \texttt{ext}: tasks available in the work of the BigScience Architecture and Scaling group \cite{scao2022language}.
\end{itemize}

When comparing with models from the state-of-the-art, we source results from a few different papers, detailed in \cref{tab:agg_sources}.

\begin{table}[h]
    \centering
    \caption{\textbf{We source evaluation results from a variety of papers across the literature, maximizing task coverage.} Although most results come from the EAI Evaluation Harness \cite{gao2021eval}, results from PaLM and GPT-3 are sourced from their respective papers. Note in Figure \ref{fig:main_lead} that the results from the GPT-3 paper are still ahead of results obtained through the API with the EAI evaluation harness.}
    \vspace{0.1in}
    \label{tab:agg_sources}
    \begin{tabular}{cccc}
    \toprule
    \textbf{Models} & \textbf{Aggregates reported} & \textbf{Source of results} & \textbf{EAI eval harness?} \\\midrule
    Ours & \texttt{main}, \texttt{core}, \texttt{ext} & This paper & $\checkmark$ \\
    BS-A\&S$^*$ & \texttt{main}, \texttt{core} & \citet{scao2022language} & $\checkmark$ \\
    GPT-Neo$^*$ & \texttt{main}, \texttt{core} & \citet{scao2022language} & $\checkmark$ \\
    PaLM$^\dagger$ & \texttt{main} & \citet{chowdhery2022palm} & \\
    GPT-3 API$^*$ & \texttt{main}, \texttt{core} & \citet{scao2022language} & $\checkmark$ \\
    GPT-3$^\dagger$ & \texttt{main} & \citet{brown2020language} & \\
    Aleph Alpha$^*$ & \texttt{core} & \citet{alephalpha} & $\checkmark$ \\
    Cerebras-GPT$^*$ & \texttt{core} & \citet{dey2023cerebras} & $\checkmark$ \\
    FairSeq$^*$ & \texttt{core} & \citet{black2022gpt} & $\checkmark$ \\
    Pythia(-Dedup)$^*$ & \texttt{core} & \citet{dey2023cerebras} & $\checkmark$ \\
    OPT$^*$ & \texttt{core} & \citet{dey2023cerebras} & $\checkmark$ \\
    GPT-J$^*$ & \texttt{core} & \citet{black2022gpt} & $\checkmark$ \\
    GPT-NeoX 20B$^*$ & \texttt{core} & \citet{black2022gpt} & $\checkmark$ \\ \bottomrule
    \end{tabular}
\end{table}

\subsection{Models}
\label{sec:other_models}

We compare against nearly 50 models across 10 series trained on a variety of curated corpora, presented in \cref{tab:models}.

\begin{figure}[h]
\centering   
\includegraphics[width=0.5\textwidth]{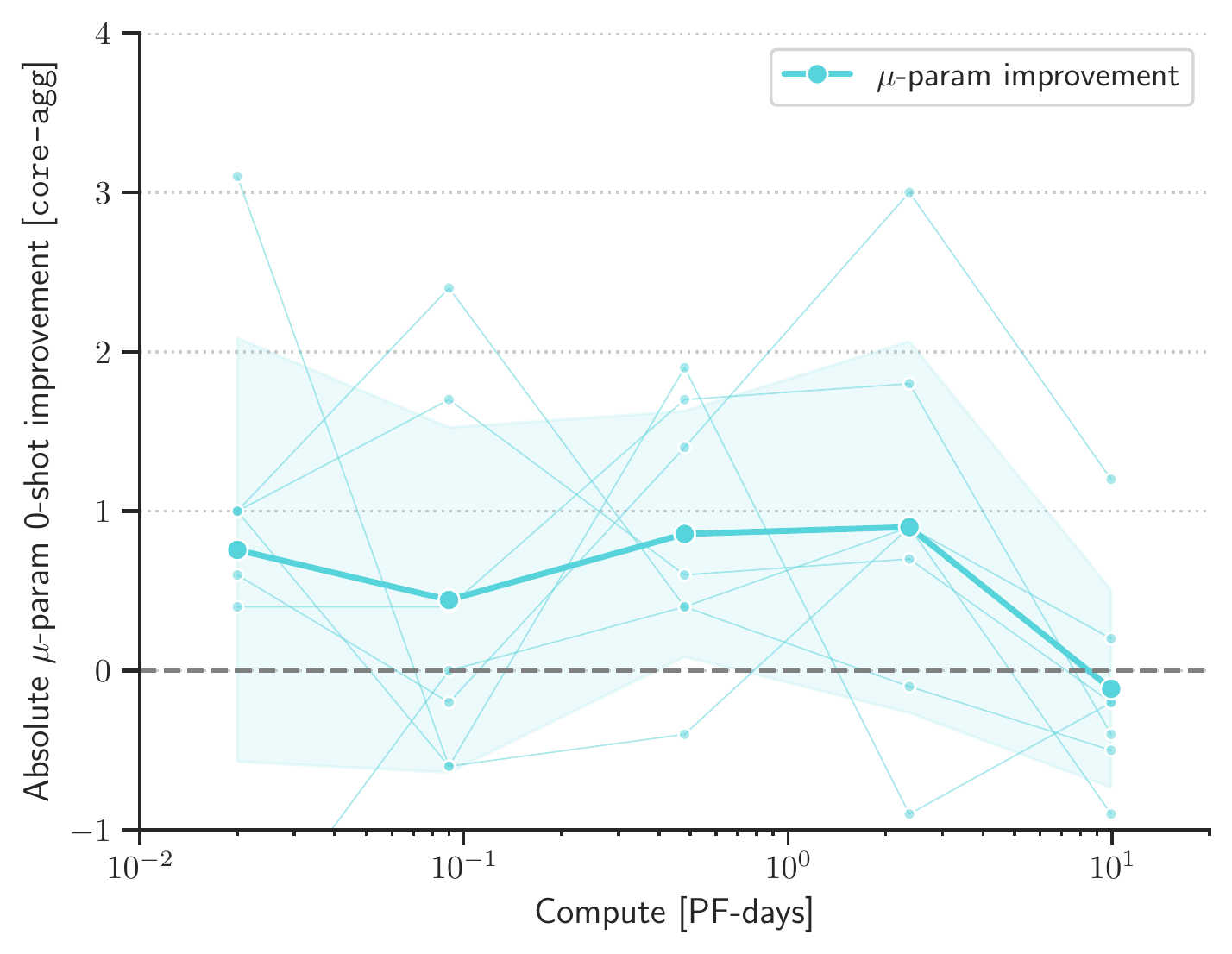}
\caption{\textbf{$\mu$-parametrization \cite{yang2021tuning} slightly improves performance in the Cerebras-GPT series \cite{dey2023cerebras}.} Zero-shot performance on our \texttt{core} aggregate, gap between Cerebras-GPT with $\mu$-param and without. Individual curves for per-task results and 1-$\sigma$ standard deviation across all tasks in the aggregate in transparent.}
\label{fig:cerebras_u}
\end{figure}

\paragraph{Cerebras-GPT with $\mu$-parametrization.} The Cerebras-GPT series \cite{dey2023cerebras} also comes in a smaller series, up to 2.7B parameters, following the recommendations of $\mu$-parametrization \cite{yang2021tuning}. As we found the performance of this smaller series to be close to the main series of models (see \cref{fig:cerebras_u}), and as it does not include models of a similar compute scale as the ones we compare to, we chose not to report it in our main figures.

\begin{figure}[h]
\centering   

\includegraphics[width=0.5\textwidth]{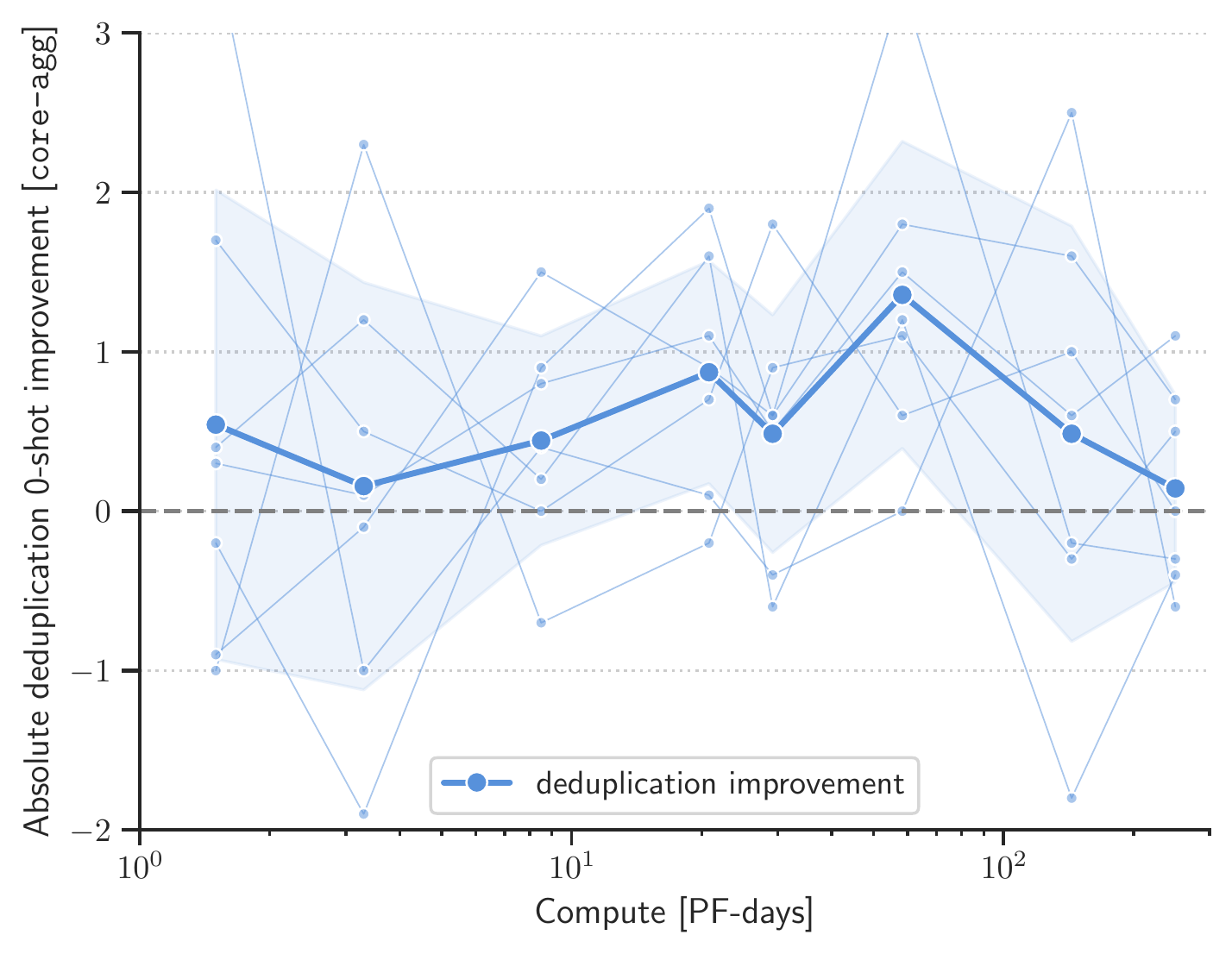}
\caption{\textbf{In our \texttt{core} aggregate, deduplication brings a small improvement to the Pythia suite \cite{biderman2023pythia}.} Zero-shot performance on our \texttt{core} aggregate, gap between Pythia trained on the deduplicated and vanilla Pile. Individual curves for per-task results and 1-$\sigma$ standard deviation across all tasks in the aggregate in transparent.}
\label{fig:pythia-dedup}
\end{figure}

\paragraph{Pythia and deduplication.} The Pythia series of models is available in two flavours: one trained on the vanilla version of The Pile, and another trained on a version deduplicated with MinHash. Performance between these two flavours was noted to minimally differ \cite{biderman2023pythia}; in \cref{fig:pythia-dedup}, we find the deduplicated version may be slightly ahead of the non-deduplicated one under our aggregate. The higher end of this improvement is broadly in line with our findings in \cref{tab:other_mdr}. Nevertheless, a difference in our findings and theirs remain. We posit a few possible hypotheses: 
\begin{itemize}
    \item \textbf{Differences between curated and web data.} It is possible that web data is more sensitive to duplicates. For instance, the most common duplicates in web data (e.g., spam) may be more detrimental than the most common duplicates in curated data. This suggests a qualitative component to deduplication that we have not studied in this work.
    \item \textbf{Differences in deduplication pipeline.} Because \citet{biderman2023pythia} uses the MinHash settings from \citet{lee2022deduplicating}, they are mostly identical to ours. However, we also apply exact deduplication: while their deduplication incurs a 30\% reduction in size, our deduplication is more aggressive, resulting in a 45\% reduction in size. This may explain why our results in \cref{tab:other_mdr} show a stronger gain from deduplication than theirs in \cref{fig:pythia-dedup}.
    \item \textbf{Differences in pretraining.} Finally, we note that \citet{biderman2023pythia} chooses to perform a partial extra epoch on the deduplicated data to reach 300GT, while we always perform a single epoch. Their setting corresponds to a data-constrained scenario, which is more realistic for the curated data they study; for us, web data is plentiful, so deduplication never truly limits the size of the datasets we can use.
\end{itemize}

\begin{table*}[h]
    \centering
    \scriptsize
    \caption{\textbf{Full-scale models trained on RefinedWeb (Falcon-RW) and other models from the state-of-the-art.} Across models trained on The Pile, the Pythia models are the closest to our achitecture: they use FlashAttention with rotary embeddings--with for only notably exception the use of parallel attention and feedforward for their models. Training budget $C$ in PF-days calculated using $C=6ND$, with $N$ the number of parameters, and $D$ the pretraining dataset size \cite{kaplan2020scaling}.}
    \label{tab:models}
    \vspace{0.1in}
    \begin{tabular}{lcccccccccccc}
    \toprule
    \textbf{Series} & \multicolumn{2}{c}{GPT-3 (paper)$^\dagger$} & \multicolumn{2}{c}{GPT-3 (API)$^*$} & \multicolumn{1}{c}{BigScience$^*$} & \multicolumn{1}{c}{PaLM$^\dagger$} & \multicolumn{3}{c}{Ours} \\\midrule
      \textbf{Model} & XL & XXL & \texttt{babbage} & \texttt{curie} & BS-A\&S & PaLM-8B & Ours (Pile) & \multicolumn{2}{c}{Falcon-RW}  \\
      \textbf{Dataset} & GPT-3 & GPT-3 & GPT-3 & GPT-3 & Pile & PaLM & Pile & RW & RW  \\
      \textbf{Params.}  & 1.3B & 6.7B & 1.3B & 6.7B & 1.3B & 8.6B & 1.3B & 1.3B & 7.5B  \\
      \textbf{Pretraining} & 300GT & 300GT & 300GT & 300GT & 300GT & 780GT & 350GT & 350GT & 350GT \\
      \textbf{PF-days} & 27 & 140 & 27 & 140 & 27 & 466 & 32 & 32 & 182\\
      \textbf{Citation} & \multicolumn{4}{c}{\citet{brown2020language}} & \multicolumn{1}{c}{\citet{scao2022language}} & \multicolumn{1}{c}{\citet{chowdhery2022palm}} & \multicolumn{3}{c}{This paper} \\ \bottomrule
    \end{tabular}
    
    \vspace{0.1in}
    
    \begin{tabular}{lcccccccccccc}
    \toprule
    \textbf{Series} & \multicolumn{3}{c}{EleutherAI$^*$} & Pythia$^*$  \\\midrule
      \textbf{Model} & GPT-Neo & GPT-J & GPT-NeoX 20B & Pythia(-Dedup)  \\
      \textbf{Dataset} & Pile & Pile & Pile & Pile (dedup)  \\
      \textbf{Params.}  & 1.3B & 6.7B & 20B & 70M-12B  \\
      \textbf{Pretraining} & 380GT & 402GT & 472GT & 300GT \\
      \textbf{PF-days} & 34 & 187 & 656 & 1.5 - 250 \\
      \textbf{Citation} & \citet{gpt-neo}& \citet{gpt-j} & \citet{black2022gpt} & \citet{biderman2023pythia}  \\ \bottomrule
    \end{tabular}

    \vspace{0.1in}

        \begin{tabular}{lcccccccccccc}
    \toprule
    \textbf{Series} &  Aleph Alpha$^*$  & Cerebras-GPT$^*$ & OPT$^*$ & FairSeq$^*$ \\\midrule
      \textbf{Model} & Luminous & Cerebras-GPT & OPT & FairSeq \\
      \textbf{Dataset} & \emph{undisclosed} & Pile & Pile (subset) + curated & curated \\
      \textbf{Params.}  & 13B & 111M-13B & 125M - 175B & 1.3 - 13B \\
      \textbf{Pretraining} & 400GT & 2 - 257GT & 300GT & 300GT \\
      \textbf{PF-days} & 361 & 0.02 - 232 & 3 - 3646 & 27 - 271\\
      \textbf{Citation} & \citet{alephalpha} & \citet{dey2023cerebras} & \citet{zhang2022opt} & \citet{artetxe2021efficient} \\ \bottomrule
    \end{tabular}

\end{table*}

\subsection{Datasets}
\label{sec:other_datasets}
We extend on \cref{tab:partial_datasets} in \cref{tab:full_datasets}, providing details on the filtering and deduplication strategies used across the litterature. 

\begin{landscape}
\begin{table*}
\centering
\caption{\textbf{Common massive web-scrape and LLM English datasets.} Datasets such as OSCAR and C4 also have significant multilingual versions, which have enjoyed wide adoption~\cite{xue2021mt5}. For OSCAR, the size corresponds to the non-deduplicated version, and is estimated from the number of words x0,75 (average number of words per tokens). }
\label{tab:full_datasets}
\begin{small}
\begin{tabular}{p{2cm}p{2cm}cccp{2.2cm}p{2cm}p{2cm}p{2.2cm}p{2cm}}
\toprule
\multicolumn{4}{l}{\textbf{General information}} & \multicolumn{6}{l}{\textbf{Web data}} \\
Dataset & Notable models & Size & Availability & Web & HTML extraction & Language ID & Heuristics & Content filtering & Deduplication \\ 
\midrule
\multicolumn{10}{c}{\textsc{\textbf{Massive web datasets}}} \\ \midrule
\textbf{C4} \cite{2020t5} & T5 \cite{2020t5} & $\sim 360$GT & Public & $100$\% & .WET files & Document-level w/ \texttt{langdetect} & Document and line-level & Rules-based: code, NSFW & \textbf{Exact}: three sentences span\\
\textbf{OSCAR 21.09} \cite{OrtizSuarezSagotRomary2019} &  & $\sim 370$GT & Public & $100$\% & .WET files & Line-level w/ fastText \cite{joulin2016fasttext} & Line $<100$ characters & None & (optional) \textbf{Exact}: per line ($\sim 55\%$ removed)\\
\textbf{OSCAR 22.01} \cite{2022arXiv220106642A} &  & $\sim 283$GT & Public & $100$\% & .WET files & Document-level w/ fastText \cite{joulin2016fasttext} & Line-level, optional document-level & Optional NSFW blocklist & (optional) \textbf{Exact}: per line\\
\midrule
\multicolumn{10}{c}{\textsc{\textbf{Curated datasets}}} \\ \midrule
\multicolumn{2}{l}{\textbf{\textcolor{openai}{$\blacksquare$ GPT-3}} \cite{brown2020language}}  & $300$GT & Private & $60$\% & Unknown & Unknown & Unknown & fastText trained on HQ-data & \textbf{Fuzzy}: minhash with 10 hashes ($\sim 10\%$ removed)\\
\textbf{\textcolor{pile}{$\blacktriangledown$ The Pile}} \cite{gao2020pile} & GPT-J \cite{gpt-j}, GPT-NeoX-20B \cite{black2022gpt}, Pythia \cite{biderman2023pythia}, Cerebras-GPT \cite{dey2023cerebras} & $\sim 340$GT & Public & $18$\% & \texttt{jusText} \cite{pomikalek2011justext} & Document-level w/ \texttt{pycld2} \cite{sites2013compact} & None & fastText on curated crawl & \textbf{Fuzzy}: minhash with 10 hashes, sim. treshold 0.5 ($\sim 26\%$ removed)\\
\textbf{MassiveWeb} \cite{gopher} & Gopher \cite{gopher}, Chinchilla \cite{hoffmann2022training} & $1,400$GT & Private & $48$\% & Custom & Unknown & Document-level & SafeSearch & \textbf{Exact \& fuzzy}: exact documents, minhash w/ sim. treshold 0.8\\
\multicolumn{2}{l}{\textbf{\textcolor{palm}{$\bigstar$ PaLM}} \cite{chowdhery2022palm}} & $780$GT & Private & $27$\% & Unknown & Unknown & Document-level & ML-based filter on HQ data & Unknown\\
\midrule
\multicolumn{10}{c}{\textsc{\textbf{Ours}}} \\ \midrule
\textcolor{refinedweb}{\CIRCLE \textbf{\textsc{RefinedWeb}}} & Falcon-RW & 5,000GT & 600GT Public & $100\%$ & \texttt{trafilatura} \cite{barbaresi-2021-trafilatura} & From CCNet \cite{wenzek2020ccnet} & Document and line-level & URL blocklist & \textbf{Exact \& fuzzy}   \\
\bottomrule
\end{tabular}
\end{small}
\end{table*}
\end{landscape}

\section{Details of the Macrodata Refinement pipeline}

\subsection{URL filtering}
\label{sec:url_details}
As discussed in \cref{sec:URLFiltering}, we base our filtering of adult documents only on the URL itself, and not on the content of the documents. This design choice was motivated by: (1) challenges in avoiding overfiltering content from minorities when using ML-based classifiers on the content of documents \cite{welbl2021challenges}; (2) NSFW words block-list applied on content~(such as the one used in C4) also resulting in overfiltering of legal and medical content \cite{dodge2021documenting}.  

Our URL filtering focuses on finding domains that are related to adult content, that may be harmful to users, or that are very likely to contain mostly unstructured text/spam (e.g., file hosting websites).
First, we aggregated a list of 4.6M domains, detailed in \cref{sec:url_blocklist_details}, that we explicitly ban; then, we built a simple URL scoring system, based on matching subwords in the URL against a list of words we curated (see \cref{sec:url_score_details}). We curated this list of words based on manual inspection, cross-referencing results with pages surfaced by ToxicBERT as being outliers in toxicity \cite{Detoxify}.

\subsubsection{URL Blocklist}
\label{sec:url_blocklist_details}

\paragraph{Origin of the list.} We use an aggregated list\footnote{\url{https://dsi.ut-capitole.fr/blacklists/}} of about 4.6M URLs that we explicitly ban. This list is broken in categories (e.g. pornography, gambling); we outline the categories we selected in \cref{tab:blacklist_caterogies}. The list is regularly updated, with an original intended usage as a blocklist for universities.

\paragraph{Curation.} We noticed the list blocked a number of domains inappropriately; while these domains were few ($<$100), they accounted for a significant portion of the data filtered by the list, as these were rather prolific domains, with thousands of pages of content. To identify these false positive domains, we applied the blocklist to a subset of 832M pages. 6.04M ($0.73\%$) pages matched with the blocklist, and the number of occurrences per URL ranged from 1 to 79k. We manually inspected all URLs matched more than 4k times, which represented an appreciable portion of the dataset. We found a number of benign domains, such as pop culture news websites, or blogging platforms, which we removed from the list.
\vspace{-0.1in}
\begin{table}[h]
\caption{\textbf{We select categories likely to contain adult or malicious content, as well as spam or unstructured text.}}
\label{tab:blacklist_caterogies}
\centering
\vspace{0.1in}
\begin{tabular}{lll}
\toprule
\textbf{Category}    & \textbf{Description}                                      & \textbf{Number of links} \\ \midrule
adult       & adult websites: from eroticism to hard pornography  & 4516478         \\
phishing    & phishing websites, malwares, etc.                & 42445           \\
dating      & dating websites                                  & 3829            \\
gambling    & online casino                                    & 1365            \\
filehosting & websites hosting files, videos, pictures, music & 909             \\
ddos        & websites related to ddos attacks                 & 421             \\
agressif    & hate, racism, etc                                & 390             \\
chat        & online chat websites                             & 244             \\
mixed adult & websites with some adult content         & 153             \\
arjel       & French regulated gambling websites               & 69              \\\bottomrule
\end{tabular}
\vspace{-0.1in}
\end{table}

\subsubsection{URL Scoring with a Word-List}
\label{sec:url_score_details}
To score URLs, we used three matching patterns based on a soft, hard, and strict violation word-list:
 \begin{itemize}[noitemsep]
    \item \textbf{Strict \underline{sub}word matching}: http://foo\textcolor{red}{bann}.\textcolor{red}{edsub}-\textcolor{red}{wo}.\textcolor{red}{rd}bar.com/any/bar, matching words such as {\small \texttt{xvideos}}, {\small \texttt{groupsex}};
    \item \textbf{Hard \underline{whole} word matching}:  http://www.foo.\textcolor{orange}{bannedword}-bar.com, with words such as {\small \texttt{porn}}, {\small \texttt{xxx}}, {\small \texttt{orgy}};
    \item \textbf{Soft word\underline{s} matching}: http://www.foo.\textcolor{blue}{soft1}-bar-\textcolor{blue}{soft2}.com, with "softer" words such as {\small \texttt{sex}}, {\small \texttt{webcam}}, {\small \texttt{escort}}.
\end{itemize}

Each list is associated with a different level of severity: for the strictest one (strict subword matching), we ban any URL matching a banned word in its substrings (as fraudulent websites may attempt to escape similar recognition schemes by breaking-up adult keywords); for the hard whole word matching, we ban URLs with a whole word matching in the list; finally, a minimum of two matches are required with the soft word matching.

We curated the lists based on manual inspection of the data, informed by top hits reported by ToxicBERT. For the strict subword matching, we included words that were unequivocally related to adult content (e.g., {\small \texttt{groupsex}}). We avoided partial unclear matches (e.g., {\small \texttt{ass}}), that may be part of neutral words (e.g., {\small \texttt{massachusetts}}). In the soft word list, we included words that do not constitute a sufficient reason to discard the document on their own, but which are suspicious when multiple words from the list result in a match. This helped with keeping medical or legal content unaffected (e.g., a single match of {\small \texttt{dick}}). 

\subsubsection{Excluded High Quality Sources}
\label{sec:excluded_sources}

Since our paper focuses on the study of RefinedWeb alone, we chose to exclude common online sources of curated data from it. This serves two objectives: (1) it strengthens our results, by ensuring that RefinedWeb doesn't end-up actually being made mostly of known high-quality sources (e.g., Wikipedia represents a significant portion of C4); (2) future works may be interested in combining RefinedWeb with existing curated copora, which would require further deduplication if they are included in RefinedWeb. Accordingly, we remove common sources used in The Pile \cite{gao2020pile} from RefinedWeb. The full list of curated data sources domains that we blocked is in Table \ref{tab:high-quality-blocked}. 

\vspace{-0.2in}
\begin{table}[h]
\caption{\textbf{RefinedWeb is stripped from common so-called high-quality sources to simplify combining it with existing curated corpora}. This blocklist is applied at the URL filtering stage, along with the adult content blocklist.\label{tab:high-quality-blocked}}
\centering
\vspace{0.1in}
\begin{tabular}{lll}
\toprule
\textbf{Curated data source}    & \textbf{Domain name blocked}       \\ \midrule
arxiv & arxiv.org \\
AskUbuntu & askubuntu.com \\ 
StackOverflow & stackoverflow.com \\
 & stackapps.com \\
 & stackexchange.com \\
 & mathoverflow.net \\
NIH Abstracts & exporter.nih.gov \\
 & ncbi.nlm.nih.gov \\
Github & github.com \\
Ubuntu IRC & irclogs.ubuntu.com \\
HackerNews & news.ycombinator.com \\
FreeLaw & courtlistener.com \\
Reddit & reddit.com \\
Europarl & statmt.org \\
United States Patents & uspto.gov \\
Wikipedia & wikipedia.org
\\\bottomrule
\end{tabular}
\end{table}

\subsection{Line-wise filtering}
\label{sec:line_details}

Despite the improvements brought forth by running text extraction with Trafilatura, we found that a number of irrelevant lines still seeped through. These lines are usually related to navigation menus, call to actions, or social media counters. Following manual inspection of the data, we devised a line-wise filtering strategy. We analyse documents line-by-line, and discard or edit the lines based on the following rules:
\begin{itemize}[noitemsep]
    \item If it is mainly composed of uppercase characters (discard);
    \item If it is only composed of numerical characters (discard);
    \item If it is a counter (e.g. {\small \texttt{3 likes}}) (discard);
    \item If it only contains one word (discard);
    \item If it is short ($\leq10$ words) and matches a pattern (edit):
    \begin{itemize}
        \item At the beginning of the line (e.g. {\small \texttt{sign-in}});
        \item At the end of the line (e.g. {\small \texttt{Read more...}});
        \item Anywhere in the line (e.g. {\small \texttt{items in cart}}).
    \end{itemize}
\end{itemize}

Finally, if the words in the flagged lines represent more than $5\%$ of the total document words, the document is discarded. We derived these filters through manual inspection of the data, and note that they require adaptation across languages.

\subsection{Deduplication}

We make use of the two deduplication methods described in \citet{lee2022deduplicating}: \textsc{ExactSubstr} and \textsc{NearDedup} (detailed in \cref{sec:minhash_details} and \cref{sec:exact_details}; see \cref{sec:dedup_samples} for samples of duplicates). 

We start with the most scalable approach, \textsc{NearDedup}. We remove similar documents by applying MinHash \citep{broder1997resemblance}, whereby a signature/sketch supporting efficient approximate similarity queries is computed for each document in the dataset, and document pairs with a high \textit{n}-gram overlap are identified. 

We then use~\textsc{ExactSubstr}, leveraging the implementation from \citet{lee2022deduplicating}\footnote{\url{https://github.com/google-research/deduplicate-text-datasets}}, to identify ranges of exact duplicate text of at least 50 tokens. We experiment with three different approaches for these ranges: \textsc{ExactSubstr-Cut}, where we remove them from the original text, as done in the original implementation; \textsc{ExactSubstr-Mask}, where the dataset is unchanged but we do not compute the loss on the duplicated ranges; and \textsc{ExactSubstr-Drop}, where we simply drop an entire document if the duplicated ranges make up more than a certain percentage of its content. 

We present small-scale ablations around these different approaches in \cref{sec:ablation_dedup}.

\subsubsection{MinHash Approximate Matching}
\label{sec:minhash_details}
We employ MinHash to find approximate duplicate documents in our web corpora at a very large scale. This technique allows us to identify templated pages or otherwise very similar content where most of the interspersed duplicated sections are small enough to not be identified by exact matching methods (anything smaller than 50 tokens).

\paragraph{Signing.} We start by normalizing the content to increase recall: punctuation is removed, text is lowercased, NFD Unicode normalization is applied, accents are removed, and all whitespace is normalized. We tokenize the resulting text using the GPT-2 tokenizer \cite{radford2019language} and obtain the set of unique \textit{n}-grams for each document. Hash functions are used to obtain a signature for each document: for each hash function, the smallest value is kept from hashing every unique \textit{n}-gram in the document. If two documents are similar, then there is a high probability that they will have the same minimum hash (MinHash) for at least some of the hash functions used \cite{broder1997resemblance}. The ratio of matching hashes between two documents approximates the Jaccard Similarity \citep{Jaccard1912THEDO} of the sets of their unique \textit{n}-grams (the sets being $d_i$ and $d_j$):

\begin{equation}
    J(d_i, d_j) = \frac{\left | d_i \cap d_j \right |}{\left | d_i \cup d_j \right |}
\end{equation}

\paragraph{Matching.} Since comparing MinHash signatures between every possible document pair is computationally expensive, we apply a locality sensitive hashing version of MinHash, MinHash LSH. A document signature is split into \textit{r} buckets, each with \textit{b} minhashes. Documents are indexed by these \textit{b} minhashes on each of the \textit{r} buckets, and we mark two documents as duplicates if their \textit{b} minhashes are exactly the same on at least one of the buckets. These two parameters, \textit{b} and \textit{r}, will determine the probability that similar documents will be detected. For two documents $i$ and $j$ whose ratio of matching hashes between their MinHash signatures is $s_{i,j}$, the probability that there is a match in a given bucket is $s_{i,j}^b$; the probability that there isn't a match in any of the buckets is $(1-s_{i,j}^b)^r$; and finally that there is a match in at least one of the buckets:

\begin{equation}
    P = 1 - (1-s_{i,j}^b)^r
\end{equation}

We use the same parameters as \citet{lee2022deduplicating}: $n=5$ (\textit{5}-grams); $b=20$ and $r=450$. This means that for each document, we compute a total of 9000 minhashes, and that the probability that a document pair with similarity 0.75 or 0.8 will be marked as duplicates will be $76\%$ and $99.4\%$ (respectively), diminishing rapidly for smaller similarity values. 

Finally, we cluster documents across all buckets --- if documents A and B match in one bucket and B and C in another, A-B-C becomes a cluster. We randomly remove all but one of the documents in each cluster.

\citet{lee2022deduplicating} also proposed filtering down on false positives by computing the real Jaccard similarity, or other metrics such as the edit similarity between identified document pairs. Given the large amount of data we have available across all of CommonCrawl, and that our main concern is improving recall, we decided to skip this additional step.

\subsubsection{Exact substring deduplication}
\label{sec:exact_details}

We make use of the \textsc{ExactSubstr} implementation publicly released by \citet{lee2022deduplicating} for exact text matching. We apply exact substring deduplication to data that has already been deduplicated by MinHash, reducing by nearly 40\% size of the dataset on which we have to operate. \textsc{ExactSubstr} will find long strings of text that are present, character for character, across multiple documents. Some of these may have escaped the earlier stage of approximate deduplication: they might not constitute a big enough portion of the document; one document might have repeated sections sourced across many different documents; or they may simply not have been found due to the approximate nature of MinHash.

\paragraph{Finding duplicates.} \textsc{ExactSubstr} concatenates all the documents in the dataset to create a single long text sequence; then, it builds a suffix array \citep{manber1993suffix} in linear time---an array of the indexes to a lexicographical ordering of all the suffixes in the sequence. Finally, duplicate sequences can also be found in linear time using the suffix array, by simply traversing the ordered list of suffixes and comparing the beginning of each pair of two consecutive suffixes. 

We apply the same normalization and tokenization as for MinHash to the content of our documents before concatenating them. One important difference is that reversibility is important: for MinHash, we were discarding entire documents, and thus never relying on the normalized+tokenized representation for downstream use. Here, once we have identified duplicate normalized+tokenized spans, we need to revert to the original span to remove it. Accordingly, we include normalization in the tokenization process, and validate that the process is reversible.

If a match is longer than 50 tokens, there will be multiple overlapping duplicated ranges. These overlapping duplicated ranges in the concatenated dataset sequence are merged before we save them to a file. We then take these ranges and retrieve the original document that produced them, obtaining the character substrings corresponding to the duplicated token ranges. 

\paragraph{Removing duplicates.} We considered applying the following transformations to the duplicate spans:

\begin{itemize}
    \item \textsc{ExactSubstr-Cut}: we remove the duplicated spans, and discard documents where there are fewer than 20 non-duplicated characters left--this is the vanilla setting used by \citet{lee2022deduplicating};
    \item \textsc{ExactSubstr-Mask}: we loss-mask the duplicated spans, preventing a loss from being computed on the duplicated text during pretraining, and discard documents where there are fewer than 20 non-masked characters left.
    \item \textsc{ExactSubstr-DropPartial}: if more than 20\% of the document is duplicated, we remove the entire document;
    \item \textsc{ExactSubstr-DropAny}: we drop any document with a duplicated span in it.
\end{itemize}

Broadly speaking, \textsc{ExactSubstr-Cut} might remove text mid-sentence resulting in disconnected text; \textsc{ExactSubstr-Mask} does not have this issue, but might be less efficient as a significant portion of the training tokens will not directly contribute to updating the model's weights; \textsc{ExactSubstr-Drop} might still keep considerable duplicated sections in its \textsc{Partial} version, especially on larger documents, while the \textsc{Any} version might be overly aggressive. Following ablations in \cref{sec:ablation_dedup}, we choose to stick with the vanilla approach, \textsc{ExactSubstr-Cut}.

Note that in all cases, while MinHash keeps one copy of the duplicated documents, our exact deduplication removes all copies of the duplicated span.

\subsection{Execution environment}
Most data processing took place in large CPU clusters, with 100-250 AWS c5.18xlarge instances; each instance has 72 vCPUs and 144 GiB of memory. We usually run with 10,000-20,000 vCPUs in the cluster, enabling rapid parallel processing.

For \textsc{ExactSubstr}, the entire dataset being deduplicated needs to be loaded onto memory: we leveraged the AWS x2iedn instances, which come with up to 2 TiB of memory in a single instance.

\newpage

\section{Deduplication samples from RefinedWeb}
\label{sec:dedup_samples}

\subsection{MinHash clusters}
\label{sec:minhash_cluster}

We report the 8 largest duplicate clusters found by MinHash in \cref{tab:minhash_clusters} -- each spanning hundreds of thousands of documents. We also found a large number of duplicate document pairs to be due to different URL GET parameters not resulting in significantly different content. An example of this behaviour can be seen in the URLs presented in \cref{tab:minhashs_examples}. 

\begin{table*}[h]
\centering
\caption{\textbf{Top-8 largest MinHash clusters found when building RefinedWeb.} We cut some of the longest samples in the interest of readability, only keeping a brief description.}
\label{tab:minhash_clusters}
\vspace{0.1in}
\begin{small}
\begin{tabular}{p{3in}|p{3.4in}}
\toprule
\textbf{Description} & \textbf{Example document} \\ 
\midrule
Wordpress sitemap notice generated by the Google Sitemap Generator Plugin
& 
This is a XML Sitemap which is supposed to be processed by search engines which follow the XML Sitemap standard like Ask.com, Bing, Google and Yahoo. It was generated using the WordPress content management system and the Google Sitemap Generator Plugin by Arne Brachhold. You can find more information about XML sitemaps on sitemaps.org and Google's list of sitemap programs. This file contains links to sub-sitemaps, follow them to see the actual sitemap content. \\ 
\midrule
Cloudflare notice to enable Javascript & 
\\
\midrule
Templated disability notice, with different phone numbers across pages & 
Welcome to our website! As we have the ability to list over one million items on our website (our selection changes all of the time), it is not feasible for a company our size to record and playback the descriptions on every item on our website. However, if you are an American with a disability we are here to help you. Please call our disability services phone line at [redacted] or [redacted] during regular business hours and one of our kind and friendly personal shoppers will help you navigate through our website, help conduct advanced searches, help you choose the item you are looking for with the specifications you are seeking, read you the specifications of any item and consult with you about the products themselves. There is no charge for the help of this personal shopper for any American with a disability. Finally, your personal shopper will explain our Privacy Policy and Terms of Service, and help you place an order if you so desire.\\
\midrule
Templated cookies notice & \\
\midrule
Templated domain name for sale page & \\
\midrule
\texttt{\small www.metoperashop.org} and sub-URLs, with content changes but always the same (large) footer & \\
\midrule
Different pages across more than 80 different domain names but with a common section & DC Customers also liked:
Special event items are produced by manufacturers only after the outcome of a game or event. These are advanced sale items and will ship immediately after they are received in our warehouse.
Manufacturer direct items are shipped directly from the manufacturer. These items are not available for international or expedited shipping.
Customized items can be personalized with options such as your name, your favorite number, and/or designs. Some options may be limited by league rules.
\\
\midrule
\texttt{\small http://www.boxofficemojo.com/daily} and sub-URLs & \\
\bottomrule
\end{tabular}
\end{small}
\end{table*}

\begin{table}[h]
\centering
\caption{\textbf{URL with different GET parameters don't always result in significantly different page content.}}
\vspace{0.1in}
\label{tab:minhashs_examples}
\begin{small}
\begin{tabular}{p{2.8in}|p{2.8in}}
\toprule
\begin{verbatim}
http://gamesandbiz.blogspot.com/2010/
07/bad-reviews-can-hurt-game-sales.ht
ml?showComment=1278486430242
\end{verbatim}
& 
\begin{verbatim}
http://gamesandbiz.blogspot.com/2010/
07/bad-reviews-can-hurt-game-sales.ht
ml?showComment=1278499674195
\end{verbatim} \\ 
\midrule
\begin{verbatim}
https://www.ocean-oxygen.org/home;jse
ssionid=1E3290E84F668552FAC643D0A8F81
BEC?p_p_id=122_INSTANCE_Zy6zjkRLAg7v&
p_p_lifecycle=0&p_p_state=normal&p_p_
mode=view&p_p_col_id=column-2&p_p_col
_pos=1&p_p_col_count=6&p_r_p_56423352
4_resetCur=true&p_r_p_564233524_categ
oryId=1346016
\end{verbatim} & 
\begin{verbatim}
https://www.ocean-oxygen.org/home?p_p
_id=122_INSTANCE_Zy6zjkRLAg7v&p_p_lif
ecycle=0&p_p_state=normal&p_p_mode=vi
ew&p_p_col_id=column-2&p_p_col_pos=1&
p_p_col_count=6&p_r_p_564233524_reset
Cur=true&p_r_p_564233524_categoryId=1
346016 
\end{verbatim}
 \\
\bottomrule
\end{tabular}
\end{small}
\end{table}

\newpage

\vspace{0.1in}

\newpage

\subsection{Exact substring matches}
\label{sec:exact_matches}

Examples of exact matches found by exact substring deduplication can be seen in Table \ref{tab:exactstr_examples}.

\begin{table}[h]
\centering
\caption{\textbf{Matches found by exact substring deduplication} (in \emph{italics}).}
\vspace{0.1in}
\label{tab:exactstr_examples}
\begin{small}
\begin{tabular}{p{2.9in}|p{2.9in}}
\toprule
it appears there is a transfer of ranking signals in this relationship. Supporting this finding is a quote from Google’s guidelines: \emph{Using JavaScript to redirect users can be a legitimate practice. For example, if you redirect users to an internal page once they’re logged in, you can use JavaScript to do so. When examining JavaScript or other redirect methods to ensure your site adheres to our guidelines, consider the intent. Keep in mind that 301 redirects are best when moving your site, but you could use a JavaScript redirect for this purpose if you don’t have access to your website’s server.} NOTE: Their experiment is based on a live page with status code 200 and NOT an inactive page. So if you want to implement this for legacy
& 
Some examples of sneaky redirects include:
- Search engines shown one type of content while users are redirected to something significantly different.
- Desktop users receive a normal page, while mobile users are redirected to a completely different spam domain. \emph{Using JavaScript to redirect users can be a legitimate practice. For example, if you redirect users to an internal page once they’re logged in, you can use JavaScript to do so. When examining JavaScript or other redirect methods to ensure your site adheres to our guidelines, consider the intent. Keep in mind that 301 redirects are best when moving your site, but you could use a JavaScript redirect for this purpose if you don’t have access to your website’s server.}\\

\midrule

Find Palm Beache FL homes for sale and other Palm Beach real estate on homesofthepalmbeaches.com. Browse and search Palm Beach houses, condos, townhomes and single-family homes by community , building, or location. \emph{Our extensive database of real estate listings provide the most comprehensive property details including home values, features and local school and neighborhood info so you can be sure that you have nearly all the facts you need upfront. Search} homesofthepalmbeaches.com today! Want a closer look at what other Palm Beach properties are available?
&
Search Stuart houses, condos, townhomes and single-family homes by price and location. \emph{Our extensive database of real estate listings provide the most comprehensive property details including home values, features and local school and neighborhood info so you can be sure that you have nearly all the facts you need upfront. Search} Stuart Listings today! Want a closer look at what other Stuart properties are available? Also search our listings for the Newest Stuart Listings and Stuart Homes with Price Reductions now.
Stuart FL Homes for Sale - Stuart Real Estate Listings FREE to search
Stuart Property\\
\midrule
\emph{To find the correct size you should measure your foot from the heel to the toe point.
Add approximately 1 - 1,5cm to get the actual inner sole length. Measure both feet and fit shoes to the larger foot.
Measure feet at the end of the day, when your feet are at their largest.} Lente shoes are women's easy slip-on leisure shoes for everyday use.
These lightweight shoes have a breathable textile mesh upper made of recycled PET bottles and cool Lycra lining.
&
\emph{To find the correct size you should measure your foot from the heel to the toe point.
Add approximately 1 - 1,5cm to get the actual inner sole length. Measure both feet and fit shoes to the larger foot.
Measure feet at the end of the day, when your feet are at their largest.} Enjoy your summer days with Masera leisure sneakers. These low-cut women's sneakers are extremely lightweight thanks to phylon midsole and breathable textile mesh upper \\
\midrule
This bandana makes the perfect addition to every fur babies birthday collection! With its sparkly crown pattern, your pup will be ready for every birthday celebration! \emph{With snaps for security, this bandana is made with love, down to the very last stitch ! 
Fabric: cotton
Care Instructions: Hand wash only, iron as needed, on low heat
Always supervise your pup while wearing Faithful Paws Co. accessories, as it could become a choking hazard if consumed.}
& 
This bandana makes the perfect addition to every fur babies summer collection! With its vibrant watercolor popsicle pattern, your pup will be ready for every summer cookout! \emph{With snaps for security, this bandana is made with love, down to the very last stitch ! 
Fabric: cotton
Care Instructions: Hand wash only, iron as needed, on low heat
Always supervise your pup while wearing Faithful Paws Co. accessories, as it could become a choking hazard if consumed.}\\
\bottomrule
\end{tabular}
\end{small}
\end{table}

\end{document}